\documentclass[10pt,journal,compsoc]{IEEEtran}
\usepackage{tikz}

\IEEEpubid{\begin{minipage}{1,1\textwidth}\ \\[12pt] {\footnotesize
  0162-8828 \copyright 2017 IEEE. Personal use is permitted, but republication/redistribution requires IEEE permission. See \url{http://www.ieee.org/publications_standards/publications/rights/index.html} for more information.}
\end{minipage}} 

\ifCLASSOPTIONcompsoc
  \usepackage[nocompress]{cite}
\else
  \usepackage{cite}
\fi
\ifCLASSINFOpdf
\else
\fi
\usepackage{filecontents}

\usepackage{graphicx}
\usepackage{amsmath,amssymb} % define this before the line numbering.
\usepackage{color}

\usepackage{algorithmicx}
\usepackage[linesnumbered,lined,boxed,commentsnumbered,ruled]{algorithm2e}

\usepackage{epsfig}
\usepackage{graphicx}
\usepackage{amsmath}
\usepackage{amssymb}
\usepackage{amsthm}

\usepackage{ragged2e}

\usepackage{indentfirst}
\usepackage{amsfonts}
\usepackage{url}
\usepackage{algpseudocode}
\def\1{\mbox{\bf 1}}
\def\A{{\bf A}}
\def\M{{\bf M}}

\def\bmu{{\boldsymbol{\mu}}}

\def\B{{\bf B}}

\def\c{{\bf c}}
\def\D{{\bf D}}

\def\I{{\bf I}}
\def\J{{\bf J}}

\def\p{{\bf p}}

\def\Q{{\bf Q}}
\def\q{{\bf q}}
\def\R{{\bf R}}
\def\Real{\mathbb{R}}

\def\s{{\bf s}}
\def\Sig{\boldsymbol{\Sigma}}
\def\U{{\bf U}}

\def\V{{\bf V}}
\def\x{{\bf x}}
\def\X{{\bf X}}

\def\Y{{\bf Y}}

\def\T{\mathcal{T}}
\def\oO{\mathcal{O}}
\theoremstyle{plain}

\include{continuous_regression.bib}

\begin{document}

%\copyrightnotice

%\title{Continuous Regression on Mutually Dependent Variables for Facial Landmark Tracking}
\title{A Functional Regression approach to Facial Landmark Tracking}

\author{Enrique~S\'anchez-Lozano, Georgios~Tzimiropoulos*, Brais Martinez*,
        \\ Fernando De la Torre and Michel~Valstar% <-this % stops a space
\IEEEcompsocitemizethanks{\IEEEcompsocthanksitem E. S\'anchez-Lozano, G. Tzimiropoulos, B.Martinez and M. Valstar are with the School of Computer Science. University of Nottingham.\protect\\
E-mail: Enrique.SanchezLozano@nottingham.ac.uk 
\IEEEcompsocthanksitem F. De la Torre is with the Robotics Institute. Carnegie Mellon University.
\IEEEcompsocthanksitem *Both authors contributed equally 
}% <-this % stops an unwanted space
\thanks{Manuscript received April 19, 2005; revised August 26, 2015.}}


\markboth{Journal of \LaTeX\ Class Files,~Vol.~14, No.~8, August~2015}%
{Shell \MakeLowercase{\textit{et al.}}: Bare Demo of IEEEtran.cls for Computer Society Journals}



\IEEEtitleabstractindextext{%
\begin{justify}
\begin{abstract}
\noindent
Linear regression is a fundamental building block in many face detection and tracking algorithms, typically used to predict shape displacements from image features through a linear mapping. This paper presents a Functional Regression solution to the least squares problem, which we coin Continuous Regression, resulting in the first real-time incremental face tracker. Contrary to prior work in Functional Regression, in which B-splines or Fourier series were used, we propose to approximate the input space by its first-order Taylor expansion, yielding a closed-form solution for the continuous domain of displacements. We then extend the continuous least squares problem to correlated variables, and demonstrate the generalisation of our approach. We incorporate Continuous Regression into the cascaded regression framework, and show its computational benefits for both training and testing. We then present a fast approach for incremental learning within Cascaded Continuous Regression, coined iCCR, and show that its complexity allows real-time face tracking, being 20 times faster than the state of the art. To the best of our knowledge, this is the first incremental face tracker that is shown to operate in real-time. We show that iCCR achieves state-of-the-art performance on the 300-VW dataset, the most recent, large-scale benchmark for face tracking.
\end{abstract}
\end{justify}

}

\maketitle

%\IEEEdisplaynontitleabstractindextext

% \fi
%\IEEEpeerreviewmaketitle

\IEEEraisesectionheading{\section{Introduction}\label{sec:introduction}}
%\section{Introduction}
%\label

\begin{figure*}[t!]
\begin{center}
\includegraphics[width=0.95\columnwidth]{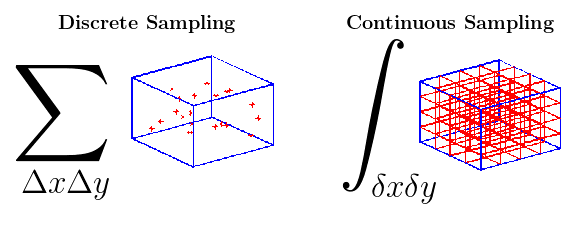} 
\vline{} 
\includegraphics[width=0.95\columnwidth]{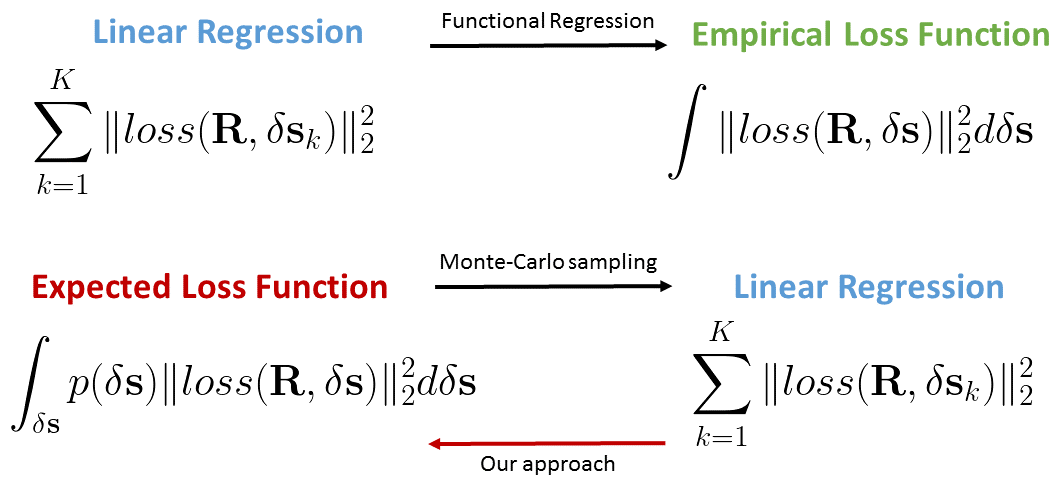}
\caption{\textbf{Left:} Difference between sampling-based regression and continuous regression. The continuous regression accounts for all the samples within a neighbourhood, whereas sampling-based needs to sample the data from a given distribution. \textbf{Right:} Our new approach to Continuous Regression can be seen as the inverse of a Monte Carlo sampling estimation. We will see that the probability density function defined by $p( \delta \s)$ defines the volume within which samples are taken.}
\label{f:abstract}
\end{center}
\end{figure*}

\IEEEPARstart{L}{inear Regression} is a standard tool in many Computer Vision problems, such as pose estimation \cite{dollar10}, and object tracking \cite{wang15b}, and it is the foremost approach for facial landmark detection, especially after the introduction of the so-called Cascaded Regression method~\cite{xiong13,xiong15,tresadern12,cootes01,tzimiropoulos15}. The goal of facial landmark detection is to locate a sparse set of facial landmarks in still images or videos. It is a problem of wide interest to the Computer Vision community because many higher level face analysis tasks, such as facial expression recognition \cite{dhall14}, and face recognition \cite{zhou03}, are critically affected by the performance of facial landmark detection systems.

Arguably, the most popular method for facial landmark detection is the Supervised Descent Method (SDM) \cite{xiong13}, in which a set of Linear Regressors are used in a cascade, each approximating the average descent direction for a specific input space. SDM generates samples by randomly perturbing the training data, i.e., each regressor is learnt by minimising the least squares error on a training set of annotated images and known displacements. Many extensions to SDM have been proposed \cite{tzimiropoulos15,sanchez16,asthana14,liu14}, yet little attention has been paid to the limitations of the \textit{sampling-based approach to linear regression}, inherent to all aforementioned methods. More specifically, it has been shown that the sampling-based approach is directly related to the following three limitations: 

\begin{itemize}
\item
The least squares formulation theoretically needs an exponential number of samples with respect to the dimensionality of the input vector in order to avoid biased models. 
\item
Training an SDM model is computationally expensive as it requires the data to be sampled per cascade level, with both memory and time needed increasing dramatically with the number of perturbations. %Therefore, training models under different configurations (e.g., for a different face detector) is in most cases very costly.
\item
Incremental learning for SDM has been reported to be extremely slow \cite{xiong14,asthana14}. This limits the capacity of SDM and its extensions for learning on-line and in real-time, which for an application such as face tracking is impractical and hence highly unsatisfactory.
\end{itemize}

In order to overcome the aforementioned limitations, in this paper we propose Continuous Regression, a Functional Regression solution to the Least-Squares problem, and show its application to real-time face tracking. Contrary to SDM, the proposed Continuous Regression solution only requires the data to be sampled at the ground-truth landmarks, i.e., no sampling at perturbations is required. This way, one can sample and store the ground-truth data only once, and then train each cascade level, or even a new model under a different configuration extremely quickly. While the SDM can take hours to train, our proposed Continuous Regression can train different models in seconds, once the data is extracted. Moreover, and contrary to existing cascaded regression methods, our continuous formulation allows for real-time incremental learning, which as we demonstrate is crucial for attaining state-of-the-art tracking performance on the most challenging subset of the 300-VW dataset~\cite{shen15}.  

Continuous Regression approaches the standard least squares formulation from a continuous perspective where the target variable is treated as a continuum. Unlike existing approaches to Functional Regression, which approximate the input space by means of B-splines \cite{marx99}, Fourier Series \cite{ratliffe02}, or Radial Basis Functions \cite{quak93}, in this paper, we propose a first-order Taylor expansion of the feature space. This way, the input space becomes linear with respect to shape displacements yielding a closed-form solution. We go beyond prior work in Functional Regression, and study the case of correlated variables which naturally arises within landmark localisation and tracking. To this end, we introduce a ``data term", tasked with correlating the different dimensions over which the problem is solved.

We then extend Continuous Regression within the cascaded regression framework, and demonstrate that training our method has notable  computational benefits  compared to that of standard SDM, without sacrificing accuracy. Finally, we devise an approach for incremental learning in the continuous domain, and show that its complexity allows for real-time tracking. To the best of our knowledge, our Incremental Cascaded Continuous Regression method (iCCR), is the first cascaded regression method that incorporates real-time incremental learning capabilities. Crucially, we show the importance of incremental learning in achieving state-of-the-art results on the 300-VW dataset.

This manuscript re-formulates the Continuous Regression over uncorrelated variables, which has been presented in \cite{sanchez12}, and further extends it to the domain of correlated variables, presented in \cite{sanchez16b}. Building on \cite{sanchez12,sanchez16b}, we present a complete formulation for the Continuous Regression problem, and then provide a geometric interpretation and link to previous works. An overview of our work is depicted in Fig.~\ref{f:abstract}.

\subsection{Contributions}
\label{ssec:contributions}

\noindent Our main contributions are as follows:

\begin{itemize}
\item 
We propose a complete formulation for Continuous Regression including a closed-form solution based on a first-order Taylor expansion of the input space. Notably, prior work in Functional Regression has only considered  approximating the input space with pre-defined basis functions. 
\item 
We then go one step further in the classical Functional Regression formulation and propose a novel optimisation problem for Continuous Regression which allows the target variable to be correlated. We also analytically derive the solution to this problem. 
\item
We incorporate Continuous Regression into the Cascaded Regression framework, demonstrating its computational benefits over SDM. We coin our method Cascaded Continuous Regression (CCR).
\item 
We derive the incremental learning updates for CCR, coined iCCR. We show that iCCR is capable of real-time incremental face tracking, being also an order of magnitude less complex than previous works based on incremental SDM.
\item 
We evaluate iCCR on the 300VW data set \cite{shen15} and show the importance of incremental learning in achieving state-of-the-art performance. 
\end{itemize}

%Code for our iCCR can be found in the author's website.

\section{Related Work}
\noindent Our work applies and extends concepts from Functional Regression to the problem of face tracking. This section provides a brief description of Functional Regression methods and how they are related to our work. We follow with a review of related face alignment and tracking methods.
\subsection{Prior Work in Functional Regression Analysis}
\label{ssec:prior_freg}

\noindent \textbf{Functional Data Analysis.} Functional Data Analysis (FDA) \cite{ramsay97}  is a branch of statistics that aims to model realisations of stochastic processes as continuous functions \cite{ramsay97}. FDA assumes that observations, and/or responses, are outcomes of continuous processes \cite{morris14}, and attempts to parameterise functions by means of basis functions, rather than as a set of samples. Several basis functions have been proposed to date for different FDA domains, such as Radial Basis Functions~\cite{quak93}, Fourier Series \cite{ratliffe02}, or B-splines \cite{marx99}.

A typical use of Functional Data Analysis is in the field of Linear Regression, called \textit{Functional Regression}\cite{yao05}, in which either the responses or the observed data, or both, are modelled as continuous functions. Functional Regression on longitudinal data can be seen as an extension of Multivariate Polynomial Regression to complex time series. Typically, basis functions are used to approximate the observed data, although some elegant solutions have also been proposed to model responses as well \cite{morris14}. However, in practice it is not computationally feasible to approximate image features by means of Radial Basis Functions or Fourier Series. We propose to approximate the input features by their Taylor expansion with respect to the variable of interest (variation on shapes) to address this. To the best of our knowledge, this is the first time that this approximation is applied in the context of Functional Regression.

\textbf{FDA in Computer Vision.}  To the best of our knowledge, there are very few works that apply ideas from FDA to Computer Vision problems, and none address the problem of facial landmark localisation. The work that is perhaps closest to ours (in some of its preliminary assumptions) is  PCA applied over continuous subspaces presented by Levin and Shashua \cite{levin02}. The main idea of \cite{levin02} is to extend the covariance matrix of the input data, necessary for the computation of PCA, to all possible linear combinations of the training samples. That is, given a set of training examples (images) ${\bf a}_i$, $i = 1...d$, the covariance is defined as $Cov(W) = \frac{1}{V}\int_{{\bf a} \in W} {\bf a} {\bf a}^T d {\bf a}$, where $W$ is the set of convex combinations of training examples, and $V$ is the volume of the polytope defined by $W$. The inverse of the volume outside the integral indicates that a uniform density function when sampling the points ${\bf a} \in W$ is assumed. In this work, we will set out how one can formulate the Least-Squares problem for Continuous Regression using a covariance matrix, providing a link between our work and \cite{levin02}. Our solution, however, does not necessarily need to be formulated using a uniform distribution. %Finally, it is worth mentioning the Continuous Generalized Procrustes Analysis of \cite{igual14}, which extends Procrustes Analysis to the continuous domain. 
 
%In \cite{levin02}, the solution is given by means of the manifold of convex combinations in an elegant manner. Their solution was shown to extend PCA for handling changes in illumination. The use of a uniform distribution is based on the fact that this way the limits are bounded, and the integral can be solved. For a different sampling distribution, the solution is not straightforward. We will show how to extend the proposed Continuous Regression to general sampling distributions in Section \ref{sec:mdcont}. 

\subsection{Prior Work on Face Tracking}

\noindent Facial landmark tracking methods have often been adaptations of facial landmark detection methods. For example, Active Shape Models \cite{cootes95}, Active Appearance Models (AAM) \cite{cootes01,matthews04}, Constrained Local Models (CLM) \cite{saragih11}, or the Supervised Descent Method (SDM) \cite{xiong13} were all presented as detection algorithms. It is thus natural to group facial landmark tracking algorithms in the same way as detection algorithms, i.e. into discriminative and generative methods. 

On the generative side, AAMs have often been used for tracking. Because  model fitting relies on gradient descent, it suffices to start the fitting from the previously tracked shape\footnote{``Implementation tricks'' can be found in \cite{tresadern12}, which provides a very detailed account of how to optimise an AAM tracker}.  AAMs have been particularly regarded as very reliable for person specific tracking, but not for generic tracking (i.e., tracking faces unseen during training) \cite{gross05}. Recently \cite{tzimiropoulos13,tzimiropoulos16} showed however that an improved optimisation procedure and the use of in-the-wild images for training can lead to satisfactorily person independent AAM. Eliminating the piecewise-affine representation and adopting a part-based model led to the Gauss-Newton Deformable Part Model (GN-DPM) \cite{tzimiropoulos14} which is the state-of-the-art AAM.

Discriminative methods directly learn a mapping from image features to shape displacements, effectively eliminating the need to minimise a reconstruction error. A certain family of discriminative methods rely on training local classifier-based models of appearance, with the local responses being then constrained by a shape model \cite{cootes95,cristinacce06,saragih11}. These algorithms are typically cast into the Constrained Local Model (CLM) formulation \cite{cristinacce06, saragih11,martins16}. 

One of the simplest tools for discriminative methods is Linear Regression, which was firstly used to bypass the need of an appearance model when fitting AAMs to images \cite{cootes01}. Herein, we refer to linear regression as a sampling-based method, in which the training is performed by generating random perturbations of the target variable (the shape displacements). Despite its simplicity and its fast computation, a single regressor has been shown to be a poor choice to account for all input variance. In order to overcome this limitation, boosted regression methods were proposed in \cite{tresadern10,cootes12} where a set of weak regressors are added to a master model, resulting in stronger regressors \cite{cootes12,martinez13}. Furthermore, in order to reduce the input variance of each of the weak regressors, \cite{dollar10} proposed the use of weakly invariant features. The use of such features requires boosted regression be split into different stages, in which a different regressor has to be learnt for each feature extraction step. This novel approach was firstly exploited for the task of pose estimation \cite{dollar10} and then was successfully applied to face alignment in \cite{cao12,cao14}. However, the most successful form of Cascaded Regression is the Supervised Descent Method (SDM) \cite{xiong13}, which applies a sampling-based linear regression for each cascade level, and in which each level can be seen as an average descent direction for the full shape displacement estimation. The use of SIFT \cite{lowe04} features, and the direct estimation of the full shape variation, made the SDM the state-of-the-art method for face alignment. Successive works have further shown impressive efficiency \cite{ren14,kazemi14} and reliable performance \cite{yan13,tzimiropoulos15}. 

However, how to best exploit discriminative cascaded regression for tracking and, in particular, how to efficiently integrate incremental learning, is still an open problem. More importantly, SDM has not yet overcome one of the main intrinsic limitations of sampling-based linear regression, to wit, the generation of biased models. This is especially important when it comes to tracking, since the initialisation is given by the points of the previous frame, and therefore there is no guarantee that the output of the model will serve as a good initialisation point for the subsequent frame. Eventually, this might cause the tracker to drift. Also, the SDM is prone to fail for large out-of-plane head poses. In order to overcome this limitation, recent methods have  proposed to pay special attention to the way the fitting is initialised. In \cite{xiong15}, a Global Supervised Descent Method applies a partition to the shape space, in order to generate a different cascaded regression model for each possible domain of homogeneous descent directions.  

Regarding evaluation and benchmarking, the only large-scale face tracking benchmark that exists to date is the 300 Videos in the Wild (300-VW) challenge \cite{shen15}. Two methods were shown to clearly outperform all others. In \cite{yang15}, a multi-view cascaded regression is used in which a different model is trained for each different partition of the space of poses. This approach, along with an accurate re-initialisation system achieved the best results in the challenge. In \cite{xiao15}, a multi-stage regression approach is used, which guides the initialisation process based on the localisation of a subset of landmarks with a strong semantic meaning. The resulting method achieved results very close to those of \cite{yang15}.

Finally, it is worth mentioning the advent of facial landmark localisation using Deep Learning \cite{xiao16,bulat16,peng16}. However, despite achieving impressive accuracy, these techniques are far from being able to operate at real-time speed.

%%% - NOTATION
\section{Linear Regression for Face Alignment}
\label{sec:lreg}
\noindent This section reviews Cascaded Regression training using Linear Regression, which is the learning problem Continuous Regression builds upon. In the following, we will represent a face image by $\I$. A face shape $\s \in \Real^{2n}$ is a vector describing the location of the $n$ landmarks considered. We will represent vector shapes as $\s = (x_1, \dots , x_n, y_1, \dots y_n)^T$. We also define $\x = f( \I , \s ) \in \Real^d $ as the feature vector representing shape $\s$. An asterisk represents the ground truth, e.g., $\s_j^*$ is the ground truth shape for image $j$. 

Cascaded Regression \cite{dollar10,cao12,xiong13} is an iterative regression method in which the output of regression at level $l$ is used as input for level $l+1$, and each level uses image features depending on the current shape estimate. The most widely used form of Cascaded Regression is the Supervised Descent Method (SDM, \cite{xiong13}), which learns a cascade of linear regressors on SIFT features. 

Training SDM for face alignment requires a set of $M$ images $\{\I_j\}_{j=1\dots M}$ and their corresponding $K$ perturbed shapes $\{\s_{j,k}^{(0)}\}_{j=1\dots M, k = 1\dots K}$. When the fitting is carried out from the output of a face detector, the different perturbations, $k = 1\dots K$, are generated by applying a random perturbation to the bounding boxes, capturing the variance of the face detector. When the fitting is initialised from the fitting of the previous frame, as is the case of tracking, the training perturbations are generated by applying a random perturbation to the \textit{ground-truth} points, ideally capturing the variance of facial shapes between consecutive frames. For each cascade level $l$ (starting with $l = 0$), a regressor is trained and applied to the training data, resulting in an update to the input shapes. Therefore, a new set of training shapes $\{\s_{j,k}^{(l+1)}\}_{j=1\dots M, k = 1\dots K}$ is generated for level $l+1$, in which the updated perturbed shapes are on average closer to their corresponding ground-truth. This way, the input variance and the training error are decreased as one descends through the cascade. 

Mathematically speaking, if we are given a set of perturbations $\delta \s^{(l)} _{j,k} = \s^{(l)}_{j,k} - \s_j^*$ between the initial shapes for level $l$, and their corresponding ground-truth, a regressor $\R^{(l)}$ is learnt that aims to minimise the least-squares error:
\begin{equation}
\label{eq:standard_LinearReg_problem}
\underset{\R}{\arg \min} \sum_{j=1}^M \sum_{k=1}^K \| \delta \s_{j,k}^{(l)} - \R f( \I_j , \s^*_j + \delta \s_{j,k}^{(l)} ) \|_2^2,
\end{equation}
where the bias term is implicitly included by appending a 1 to the feature vector. If we store the features $\x_{j,k} = f( \I_j, \s^{(l)}_{j,k})$ into matrix $\X^{(l)}$, and the corresponding displacements $\delta \s^{(l)} _{j,k} = \s^{(l)}_{j,k} - \s_j^*$ into matrix $\Y^{(l)}$, the solution to Eq.~(\ref{eq:standard_LinearReg_problem}) can be written in closed form:
\begin{equation}
\label{eq:linear_reg}
\R^{(l)} = \Y^{(l)} {(\X^{(l)})}^T \left( \X^{(l)} {(\X^{(l)})}^T \right)^{-1}.
\end{equation} 
The regressor $\R^{(l)}$ is then applied to each of the extracted features $\x_{j,k}$, and a new training set is generated, in which the training shapes are now given as:
\begin{equation}
\label{eq:linear_update}
\s^{(l+1)}_{j,k} = \s^{(l)}_{j,k} - \R^{(l)} f(\I_j, \s^{(l)}_{j,k}).
\end{equation}
The process is repeated until the average of the differences $\delta \s^{(l)}_{j,k} = \s^{(l)}_{j,k} - \s_j^*$ no longer decreases. Typically, the training error is minimised after 4 or 5 cascade levels. %Fitting is straightforward: the input shape is consequently forwarded through the cascade. Please make this 

During testing, the initial shape $\s^{(0)}$ is forwarded to the first regressor $\R^{(0)}$, to generate a new shape $\s^{(1)}$, which is subsequently forwarded to $\R^{(1)}$. For each level $l$, a new shape $\s^{(l+1)}$ is generated by applying regressor $\R^{(l)}$ to the features extracted at $\s^{(l)}$:
\begin{equation}
\s^{(l+1)} = \s^{(l)} - \R^{(l)} f(\I,\s^{(l)}). 
\end{equation}
% please edit here 

Provided the prediction statistics of $\R^{(l)}$ are known for all cascade levels, the training of Cascaded Regression can be done independently for each of them (par-SDM, \cite{asthana14}). As pointed out in \cite{asthana14}, for a sufficiently large and general training set, we can first train SDM, following the sequential algorithm shown above. After SDM training, we can compute the mean $\bmu^{(l)}$ and covariance $\Sig^{(l)}$ of the differences that result from updating the training shapes: $\delta \s^{(l)}_{j,k} = \s^{(l)}_{j,k} - \s^*_{j}$. Then, for each cascade level, we can generate a set of random perturbations $\delta \s^{(l)}_{j,k}$, drawn from a Gaussian distribution $\mathcal{N}(\bmu^{(l)},\Sig^{(l)})$, and then compute each $\R^{(l)}$ using Eq.~(\ref{eq:linear_reg}). This way, computing each regressor is independent from the other levels, meaning that the training can be performed in parallel. 

As empirically shown by \cite{asthana14}, such a model achieves similar accuracy as a model trained in a sequential manner. In addition, \cite{asthana14} showed that one can use these statistics to train a par-SDM using a larger training set, without compromising accuracy with respect to a sequentially trained SDM. This implies that, for a sufficiently large original training set, the computed statistics properly model the displacements for each level. We will see that our Continuous Regression can be straightforwardly introduced within the Cascaded Regression framework, with the advantage that our proposed method only needs the data to be sampled once, whereas both the SDM and the par-SDM require a sampling step for each Cascade Level. 

%%% - UNCORRELATED CONTINUOUS REGRESSION
\section{Continuous Regression}
\label{sec:uncor_creg}
\noindent In this section we present a complete formulation and solution for Continuous Regression, in which the shape displacements are treated as continuous variables. We propose the use of a first-order Taylor expansion approximation of the feature space, which yields a closed-form solution  in the continuous domain for the Least-Squares problem. We will see that our Taylor-based closed-form solution depends only on the features extracted from the ground-truth shapes and their derivatives. Contrary to existing approaches, \textbf{our Continuous Regression solution does not need to sample over perturbations}.

Let us consider the Linear Regression problem shown in Eq.~(\ref{eq:standard_LinearReg_problem}). The extension of the Least-Squares problem to the continuous domain of shape displacements is defined as: 
\begin{equation}
\label{eq:iid_cr}
 \underset{\R}{\arg \min}\sum_{j=1}^M \int_{-a_1}^{a_1}  \dots \ \int_{-a_{2n}}^{a_{2n}} \|\delta \s - \R f(\I_j, \s_j^*+ \delta \s)\|_2^2 d \delta \s,
\end{equation}
where\footnote{For the sake of clarity, we drop the dependence on the level} the line element is defined as $\delta \s = \delta {x_1} \dots \delta {x_n}, \delta {y_1} \dots \delta {y_n}$,  the limits are bounded, and defined independently for each dimension. Without loss of generality, we will assume that the limits are symmetric, and the displacement assumed for each of the points is therefore unbiased. 

Unlike previous works on Functional Regression, which  approximate the input feature space by B-splines or Fourier series, we propose to approximate the input feature space by its first-order Taylor expansion:
\begin{equation}
f(\I_j,\s_j^*+\delta \s ) \approx f(\I_j,\s_j^*) + \J^*_j \delta \s,
\label{eq:taylor_feat_general}
\end{equation}
\noindent where  $\J^*_j = \frac{\partial f(\I_j, \s)}{\partial \s} \vert_{(\s = \s_j^*)} \in \Real^{d \times 2n}$, evaluated at $\s = \s_j^*$, is the Jacobian of the feature representation of image $\I_j$, with respect to shape coordinates $\s$, at $\s_j^*$.
To compute the Jacobian of the image features (pixels) with respect to the $x$-positions, we have to use some approximation, e.g. the Sobel filter. We can also compute the empirical derivative of other complex features, such as SIFT or HOG. Indeed, gradient-based features, such as SIFT or HOG, are smoother than pixels, and are therefore more feasible to be approximated by a first-order Taylor expansion. Experiments will show that the use of HOG features suffices to attain state of the art performance, reinforcing the idea that a first-order Taylor expansion is sufficient. Herein, we will define the empirical derivative of the image features with respect to the $x$ (or $y$) coordinate, as follows:
\begin{equation}
\label{eq:derivative}
\nabla f_x = \frac{\partial f( \I,x)}{\partial x} \approx \frac{f( \I, x + \Delta x ) - f( \I, x - \Delta x )}{2 \Delta x}
\end{equation}
Then, the $i$-th column of the Jacobian $\J^*$, for $i = 1, \dots, n$, is given by $\nabla f_{x_i}$, whereas the $(i+n)$-th column of the Jacobian is given by $\nabla f_{y_i}$. In practice, $\Delta x$ is set to the minimum displacement, i.e., $\Delta x = 1$. Given the approximation of Eq.~(\ref{eq:taylor_feat_general}), the original problem can be expressed as:
%\begin{eqnarray}
%\int_{a_1 \dots a_{2n}} \|\delta \s - \R f(\I_j, \s_j^*+ \delta \s)\|_2^2 d \delta \s \approx \nonumber \\
%\int \| \delta \s - \R \left( f(\I_j,\s_j^*) + \J^*_j \delta \s \right)\|_2^2 d \delta \s \nonumber = \\
%\int \Big[ \delta\s^T\delta\s - 2 \delta\s^T \R (\x^*_j + \J^*_j\delta\s) + \nonumber \\ 
%(\x^*_j + \J^*_j \delta \s)^T\R^T\R(\x^*_j + \J^*_j \delta\s) \Big] d \delta \s
%\end{eqnarray}
%We can group independent, linear, and quadratic terms with respect to $\delta \s$ :
\begin{eqnarray}
\int \| \delta \s - \R \left( f(\I_j,\s_j^*) + \J^*_j \delta \s \right)\|_2^2 d \delta \s \approx \\
\int \big[ \delta \s^T \A_j \delta \s   + 2\delta\s^T\mathbf{b}_j + {\x^*_j}^T \R^T \R \x^*_j \big] d \delta \s,  
\label{eq:long_deriv_CCR2_single}
\end{eqnarray}
where $\A_j = ( \mathbb{I} - \R \J^*_j )^T ( \mathbb{I} - \R \J^*_j )$ and $\mathbf{b}_j = {\J^*_j}^T \R^T \R \x^*_j - \R \x^*_j$. Given the independence between each dimension, the linear term equals to zero for symmetric limits. For the quadratic term, the solution stems from the fact that 
\begin{equation}
\int \delta \s^T \A_j \delta \s d \delta \s = \sum_{u,v} \int A_j^{uv} \delta s_u \delta s_v d \delta \s,
\end{equation}
where $A_j^{uv}$ is the $\{u,v\}$-th entry of $\A_j$, and $s_u$, $s_v$ refer indistinctly to any pair of elements of $\s$. We can see that 
\begin{equation}
\label{eq:cases}
\int A_j^{uv} \delta s_u \delta s_v d \delta \s = \begin{cases}
   A_j^{uu} \frac{a_u^2}{3} \prod_k {2a_k}        & \quad \text{if } u = v\\
    0  & \quad \text{otherwise}\\
  \end{cases}
\end{equation}

Let $V = \prod_k {2a_k}$, and let $\Sig \in \Real^{2n}$ be a diagonal matrix the entries of which are defined by $\frac{a_u^2}{3}$. We can see that $ \int \delta \s^T \A_j \delta \s d \delta \s = V \sum_u A_j^{uu} \frac{a_u^2}{3}$. We can further observe that $\sum_u A_j^{uu} \frac{a_u^2}{3} = Tr( \A_j \Sig )$. Similarly, we can see that $\int {\x^*_j}^T \R^T \R \x^*_j d \delta \s = {\x^*_j}^T \R^T \R \x^*_j V $. Then, the solution to Eq.~(\ref{eq:long_deriv_CCR2_single}) is given by:
\begin{equation}
\label{eq:closed_form_single}
Tr( \A_j \Sig ) V + {\x^*_j}^T \R^T \R \x^*_j V.
\end{equation}

The minimisation of Eq.~(\ref{eq:closed_form_cont_single}) w.r.t. $\R$ has a closed-form as follows: 
\begin{equation}
\label{eq:closed_form_cont_single}
\R = \Sig \left( \sum_{j=1}^M  {\J_j^*}^T \right) \left( \sum_{j=1}^M \x_j^*{\x_j^*}^T + \J_j^*\Sig{\J_j^*}^T \right) ^{-1}.
\end{equation}

The solution shown in Eq.~(\ref{eq:closed_form_cont_single}) only accounts for the ground-truth features and their corresponding Jacobians. We can see that, once the data has been sampled, training a new regressor for different chosen limits is straightforward, as it does not require recalculation of perturbations. This implies that, for the Cascaded Regression framework, we only need to sample the images once. This is a huge time saving compared to the sampling-based SDM.

%%% MD-CR
\section{Continuous Regression on Correlated Variables}
\label{sec:mdcont}
\noindent This section extends Continuous Regression to the space of correlated variables. We will show that it is possible to find a closed-form solution for a general case, by assuming a different measure~(Section~\ref{ssec:mdcont}). We then study the validity of the Taylor expansion to approximate the input features~(Section~\ref{ssec:validity}). We later introduce our solution into the Cascaded Regression framework, and show how our Cascaded Continuous Regression (CCR) implies a huge training time saving (Section~\ref{ssec:ccr}). We also study the importance of the new solution (Section~\ref{ssec:data_term}). Then, we provide a theoretical link to previous work (Section~\ref{ssec:discussion}) and a geometrical interpretation (Section~\ref{ssec:geometric}).

\subsection{Reformulating Continuous Regression}
\label{ssec:mdcont}

\noindent The main problem of existing Functional Regression approaches is that the integral is defined with respect to the Lebesgue measure, which implicitly assumes that an (unnormalised) uniform distribution, in which samples are uncorrelated, is used. Thus, it is not possible to solve it for correlated dimensions. To overcome this limitation, we need to solve the integral with respect to a different measure $\mu$. We can readily see that the proper measure is the \textit{probability measure}, i.e., $\mu = \mbox{Pr}$. Therefore, the problem becomes
\begin{equation}
\label{eq:train_problem_def_pr}
\sum_{j=1}^M \int_{\delta \s}  \| \delta \s - \R f( \I_j, \s^*_j+\delta \s ) \|_2^2 d \mbox{Pr}( \delta \s ).
\end{equation}
Applying the Riemannian form, we can write Eq.~(\ref{eq:train_problem_def_pr}) as
\begin{equation}
\label{eq:train_problem_def}
\sum_{j=1}^M \int_{\delta \s}  \| \delta \s - \R f( \I_j, \s^*_j+\delta \s ) \|_2^2 p( \delta \s ) d \delta \s,
\end{equation}
where now $p(\delta \s)$ accounts for the pdf of the sampling distribution. Interestingly, we can see that by formulating the integral with respect to a probability measure, we are actually formulating Continuous Regression by means of the average \textit{expected loss function}, rather than from the classical Functional Regression perspective, which minimises the \textit{empirical loss function}. The expected loss is actually the function from which the empirical loss is derived, for example, in \cite{xiong13}, the expected loss function is reduced to the empirical loss by applying a Monte Carlo sampling approximation. Again, we will approximate the feature space by its first-order Taylor expansion. The integrals are now generally unbounded (they become bounded for e.g. a uniform distribution). Following the steps carried out in Section~\ref{sec:uncor_creg}, we expand the features, and group linear, quadratic, and independent terms, with respect to $\delta \s$.
\begin{eqnarray}
\int_{\delta \s} p( \delta \s ) \| \delta \s - \R f( \I_j, \s^*_j+\delta\s ) \|_2^2 d \delta \s \nonumber \approx \\ 
\approx \int_{\delta \s} p( \delta \s ) \big[ \delta \s^T \A_j \delta \s   + 2\delta\s^T\mathbf{b}_j + {\x^*_j}^T \R^T \R \x^*_j \big] d \delta \s,  
\label{eq:long_deriv_CCR2}
\end{eqnarray}
where, as previously, $\A_j = ( \mathbb{I} - \R \J^*_j )^T ( \mathbb{I} - \R \J^*_j )$ and $\mathbf{b}_j = {\J^*_j}^T \R^T \R \x^*_j - \R \x^*_j$. We only require the pdf $p(\delta \s)$ to be parameterised by its mean $\bmu$ and covariance $\Sig$, so that, for any symmetric matrix $\A$, the following properties hold \cite{brookes11}:
\begin{eqnarray}
\int_{\delta \s} p(\delta \s) d\delta \s = 1,  \quad \int_{\delta \s} \delta \s p(\delta \s) d\delta \s = \bmu, \nonumber \\ 
\int_{\delta \s} p(\delta \s)\delta \s^T \A \delta \s d\delta \s = Tr(\A \Sig ) + \bmu^T\A\bmu .
\label{eq:integrals2}
\end{eqnarray}
It is then straightforward to compute the error, for the $j$-th training example:
\begin{eqnarray}
\int_{\delta \s} p( \delta \s ) \| \delta \s - \R f( \I_j, \s^*_j+\delta \s ) \|_2^2 d \delta \s  \approx  \nonumber \\ 
Tr( \A_j \Sig ) + \bmu^T \A_j \bmu + 2 \bmu^T \mathbf{b}_j + {\x^*_j}^T \R^T \R \x^*_j .
\label{eq:closed_form}
\end{eqnarray}
Again, $\R$ is obtained after minimising Eq.~(\ref{eq:closed_form}) w.r.t. $\R$, in which derivatives are obtained as follows:

\begin{eqnarray}
\frac{\partial}{\partial \R} Tr( \A_j \Sig ) &=& 2 \R \J^*_j \Sig {\J^*_j}^T - 2 \Sig {\J^*_j}^T \nonumber \\
\frac{\partial}{\partial \R} \bmu^T \A_j \bmu &=& 2 \R \J^*_j \bmu \bmu^T {\J^*_j}^T - 2\bmu \bmu^T {\J^*_j}^T\nonumber \\
\frac{\partial}{\partial \R} 2 \bmu^T \mathbf{b}_j &=& 4 \R \x_j^*\bmu^T{\J_j^*}^T  - 2\bmu {\x^*_j}^T \nonumber \\
\frac{\partial}{\partial \R} {\x^*_j}^T \R^T \R \x^*_j &=& 2 \R \x^*_j {\x^*_j}^T.
\end{eqnarray}
This leads to the closed-form solution:
\begin{multline}
\label{eq:closed_form_cont}
\R = \left( \sum_{j=1}^M  \bmu {\x_j^*}^T+ (\Sig + \bmu \bmu^T){\J_j^*}^T \right) \cdot  \\
\left( \sum_{j=1}^M \x_j^*{\x_j^*}^T+2\x_j^*\bmu^T{\J_j^*}^T + \J_j^*(\Sig+ \bmu\bmu^T){\J_j^*}^T \right) ^{-1}.
\end{multline}

The similarities between the closed form error in Eq.~(\ref{eq:closed_form}) and that of Eq.~(\ref{eq:closed_form_single}) are clear. If $p(\delta \s)$ is defined as a zero-mean uniform distribution with diagonal covariance matrix with entries defined as $\frac{a^2}{3}$ (i.e., defined for the limits $a_1,\dots,a_{2n}$),  then Eq.~(\ref{eq:closed_form_cont}) and Eq.~(\ref{eq:closed_form_cont_single}) would be the same. \\
It is worth highlighting that the solution does not depend on the actual sampling pdf, but rather on its first and second order moments, i.e. Continuous Regression bypasses the question of which distribution should be used when sampling the data. 

% please emphasize the difference with sampling based methods
The proposed formulation has important computational advantages besides the theoretical differences between Continuous Regression and sampling-based methods like SDM. In particular, once the training images have been sampled, training a new regressor under a different configuration requires very little computation as opposed to the sampling-based approach. The reason for this is that Continuous Regression does not require the sampling process to be repeated, instead it only changes the sampling statistics. For instance, in the context of Cascaded Regression, given the statistics for each level of the cascade with respect to the ground truth, there is no need to sample the images more than once, which yields a huge time saving. Furthermore, we can see that there is a matrix form that can help compute Eq.~(\ref{eq:closed_form_cont}). Let us introduce the following shorthand notation: $\M = [ \bmu , \Sig + \bmu \bmu^T ]$, ${\bf B} = \bigl( \begin{smallmatrix} 1 & \bmu^T \\ \bmu & \Sig + \bmu \bmu^T \end{smallmatrix} \bigr)$, $\D_j^* = [ \x_j^*, \J_j^*]$ and $\bar{\D}^*=\left[ \D_1^*,\ldots, \D_M^* \right] $. Then:
\begin{equation}
\label{eq:compact_form_cont}
\R = \M \left( \sum_{j=1}^{M} \D_j^* \right)^T \left(\bar{\D}^* \hat{\B} \bar{\D}^{*^T} \right)^{-1}
\end{equation}

\noindent where $\hat{\bf B} = {\bf B} \otimes {\mathbb{I}}_M$. Since $\hat{\bf B}$ is sparse, computing Eq.~(\ref{eq:compact_form_cont}), once the data is given, is done in a matter of seconds. % more discussion/comparison

\subsection{Validity of Taylor expansion}
\label{ssec:validity}
\noindent In order to validate the Taylor approximation to represent the image features, we have designed the following experiment. For a given set of initial statistics $\Sig^l$, a set of random perturbations ${f(\I,\s^* + \delta \s)}$ was created, where $\delta \s$ was drawn from $\mathcal{N}({\bf 0}, \Sig^l)$. Then, we measured the distance between the Taylor approximation to the ground-truth features, as well as to the features extracted from the same image at different locations. Comparing these two distances allows to distinguish whether the Taylor expansion gives a reasonable estimation or a random estimation. The distance between the Taylor approximation at $\delta \s_i$ and the features collected at $\delta \s_j$ (including $i = j$) is given as:
\begin{equation}
dist_{i,j} = \frac{\|f(\I,\s^* + \delta \s_j) - ( f(\I,\s^*) + f'(\I,\s^*) \delta \s_i)  \|}{ \| f(\I,\s^* + \delta \s_j) + ( f(\I,\s^*) + f'(\I,\s^*) \delta \s_i) \|},
\label{eq:1}
\end{equation}
This distance is expected to increase with $\delta \s$. However, despite this increase, we expect samples to be distinguishable from samples taken from other images. 

\begin{figure}[h!]
\begin{center}
\includegraphics[width=0.9\columnwidth]{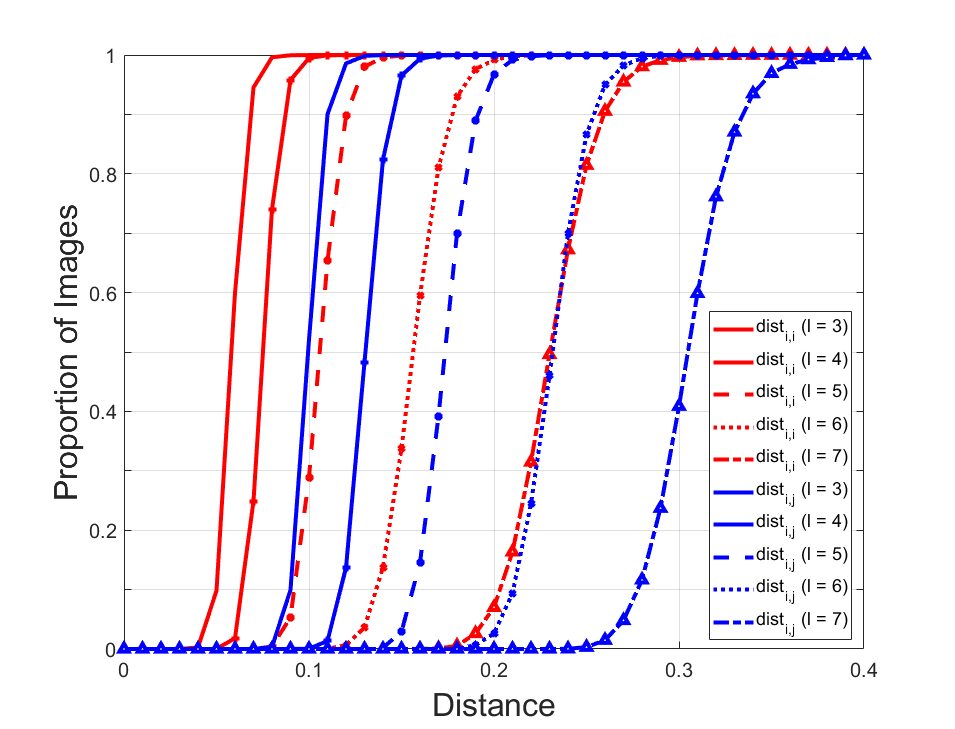}
\caption{Results obtained for the Taylor approximation. Red curves correspond to the distances described in Eq.~(\ref{eq:1}), when $i=j$, for a varying set of initial perturbations ranging from $l = 3$ to $l = 7$. Blue curves correspond to the distances described in Eq.~(\ref{eq:1}) when $i \neq j$. \label{fig_1}}
\end{center}
\end{figure}
We have used a random subset of 1000 images from the training set (see Section 7 for further details). We have evaluated the distances at the locations defined by a set of $5$ different $\Sig^l$. More specifically, for $l = 3 \dots 7$, $\Sig^{l}$ is defined as a diagonal matrix the elements of which are set to $2^{l-1}$ (which represents the variance of the displacement per landmark, in pixels). The results are shown in Fig.~\ref{fig_1}. It can be seen that even when the variance is fairly high, the approximated samples are still distinguishable from samples collected at other locations (i.e. red lines have lower error than corresponding blue lines). This validates the Taylor approximation for a far long margin. 

\begin{figure}[h!]
\begin{center}
\includegraphics[width=0.9\columnwidth]{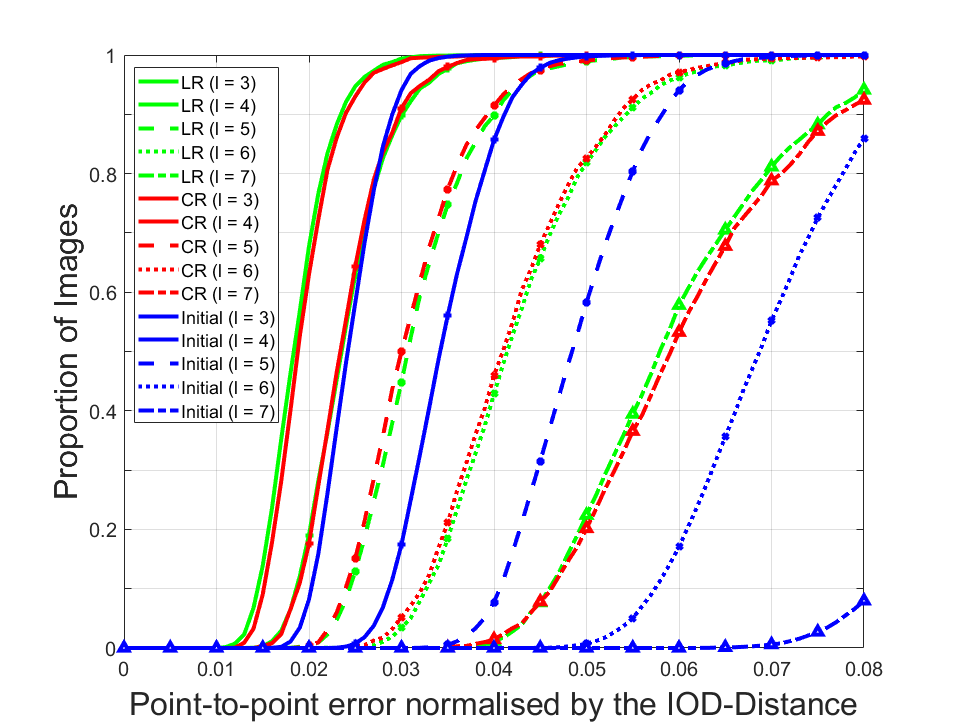}
\caption{Results attained by a sampling-based linear regression (green curves) and a regressor trained using Continuous Regression (red curves) for the given initial perturbations (blue curves) \label{fig_2}}
\end{center}
\end{figure}

In order to also check the capabilities of a regressor trained using a Taylor approximation, we have trained two regressors for each of the $\Sig^l$ described above (one using the sampling-based approach, and one using Continuous Regression). For each of the images, a new set of random locations was generated, and the regressors were used to predict the shape displacements. Results are shown in Fig.~\ref{fig_2}. It can be seen that both regressors have similar accuracy, thus validating the assumptions made through the paper. 
It has to be noted that these validations hold for such a large range thanks to the smoothness and the constrained environment that face tracking generally offers. When the displacement is far from the ground-truth, the distance between the approximation and the sampled data increases, but so it does when collecting the data somewhere else. Therefore, there is still the possibility of training a regressor capable of reducing the error, as shown in Fig.~\ref{fig_2}. 

\subsection{Cascaded Continuous Regression}
\label{ssec:ccr}
\noindent We can readily extend Continuous Regression to the Cascaded Regression framework. To do so, we point out that the statistics defined in Eq.~(\ref{eq:closed_form_cont}) correspond to those used in the parallel SDM training proposed in \cite{asthana14}, described in Section~\ref{sec:lreg}. In this context, we can train each cascade level using the corresponding $\bmu^{(l)}$ and $\Sig^{(l)}$. We coin our method Cascaded Continuous Regression (CCR). The main advantage of our CCR with respect to SDM is that we only need to sample the training images once, given that Eq.~(\ref{eq:closed_form_cont}) only needs to account for the ground-truth features and their corresponding Jacobians. That is to say, the \textbf{sampling process in the CCR training is done only once}. 

Denoting the cost of sampling an image as $\oO(q)$, for a set of $M$ images, $K$ perturbations, and $L$ cascade levels, the total sampling cost of SDM is $\oO(LKMq)$. In our proposed CCR, the sampling is done only once, which is $\oO(5qM)$, given that extracting the ground-truth features and Jacobians is $\oO(5q)$. Therefore, our CCR training presents a computational advantage, with respect to SDM, of $LK/5$. In our setting, $L = 4$ and $K = 10$, meaning that the Continuous Regression sampling is $8$ times faster (in FLOPS). However, we have to note that, if we were to train a different model \textbf{for a different set of statistics}, we would not need to sample any data again. This means that, under a different configuration, we only need Eq.~(\ref{eq:compact_form_cont}) to be computed, which can be done in seconds. In this case, we can see that the total sampling cost of SDM training would remain $\oO(LKMq)$, while retraining a CCR does not entail any sampling, thus having a null cost. In our configuration, $M \approx 7000$ images, meaning that \textbf{the computational saving of retraining a CCR, with respect to SDM, is} $\oO(10^6)$. 

In the case that no prior SDM has been trained, Algorithm~\ref{alg:ccr_train} summarises the training of CCR. Given the training images and ground-truth points, CCR needs the data to be pre-computed, by computing all $\D^*_j$. Then, the process of CCR training basically consists of updating the initial statistics $\bmu^0$ and $\Sig^0$ (computed for a training set of annotated videos), to generate a new regressor for each cascade level. The statistics for each level are generated by computing the difference of the outputs of the previous level with respect to the corresponding ground-truth points. For the first level, a set of initial shapes $\s_j^0$ are generated by randomly perturbing the ground-truth with $\bmu^0$ and $\Sig^0$. 

\begin{algorithm}[h!]
\label{alg:ccr_train}
\DontPrintSemicolon
\SetAlgoLined
\KwData{$\{\I_j,\s^*_j, \s^0_j\}_{j=1:M}$, $\bmu^0$ and $\Sig^0$; $L$ levels}
Pre-compute: Extract $\bar{\D} = \{\D^*_j\}_{j=1:M}$\;
Pre-compute: $\hat{\D} = \left( \sum_{j=1}^{M} \D_j^* \right)^T$ \;
\For{$l = 0 : L - 1$}{
Compute $\M_l$ and ${\bf B}_l$ \;
$\R_l = \M_l \hat{\D} \left(\bar{\D} \hat{\B}_l \bar{\D}^T \right)^{-1} $\;
Apply $\R_l$ to $\s^{(l)}_j$ to generate $\s^{(l+1)}_j$ \;
Compute distances $\delta \s_j = \s^{(l+1)}_j - \s_j^*$ \;
Update $\bmu^{l+1} = mean( \{\delta \s\} )$ and $\Sig^{l+1} = cov( \{\delta \s \}) $ \;
}
\caption{CCR training}
\end{algorithm}

%%%%
\subsection{The impact of the data term}
\label{ssec:data_term}
\begin{figure}
\begin{center}
\includegraphics[width=0.9\columnwidth]{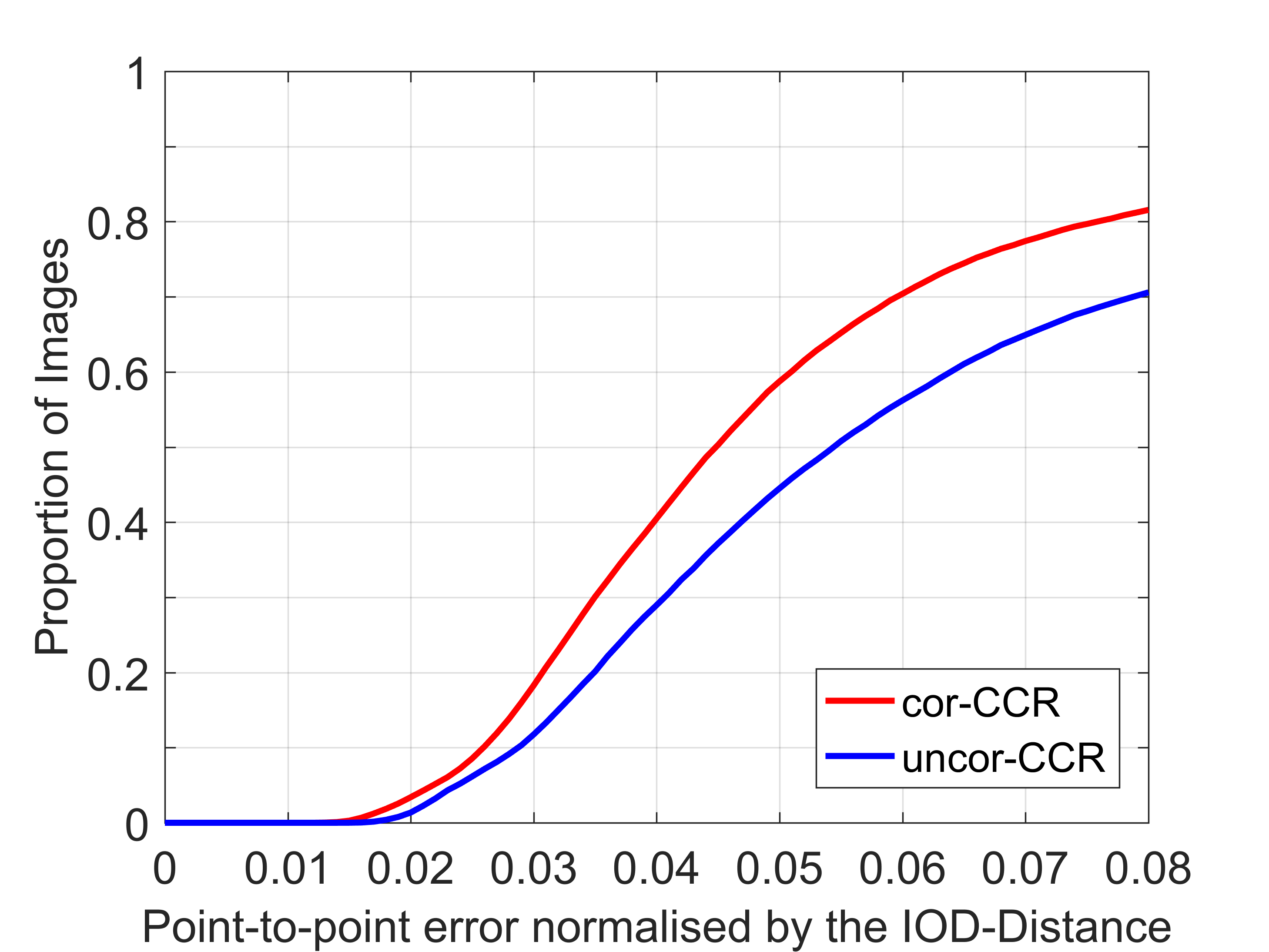}
\end{center}
\caption{Cumulative Error Distribution (CED) curve for both the uncor-CR (red) and cor-CR (blue). Results are shown for the 49 points configuration. The contribution of a full covariance matrix is clear \label{fig:dataterm}}
\end{figure}

\noindent To empirically demonstrate the theoretical influence of the data term which allows for correlated variables, we compared the performance of a CCR model trained using the ``correlated" solution in Eq.~(\ref{eq:closed_form_cont}) (cor-CCR), with a CCR model trained using the ``uncorrelated" version described in Section~\ref{sec:uncor_creg} (uncor-CCR). Both models were trained on the Helen database~\cite{le12}, using Algorithm~\ref{alg:ccr_train}. In both cases, the initial training set is generated by perturbing the ground-truth shapes according to the statistics measured at the training partition of 300-VW dataset~\cite{shen15} (see Section~\ref{sec:experimental_results} for further details). In  uncor-CCR, the statistics for each level were forced to have zero-mean and diagonal covariance. Otherwise the two versions are the same. We evaluated both models in the most challenging subset of the 300-VW test partition. Results in Fig.~\ref{fig:dataterm} show the Cumulative Error Distribution (CED). It is immediately clear that the contribution of cor-CCR is significant, and thus that shape dimensions are significantly correlated. The result of this experiment clearly demonstrates the importance of solving Continuous Regression in spaces of correlated variables. 

\subsection{Connection to related work}
\label{ssec:discussion}
\noindent In this section, we link our work to that of \cite{levin02}. More specifically, we show that one can first compute the functional covariance in a similar fashion to that of \cite{levin02}, and then link it to the Least-Squares solution. To do so, we first observe that the normal equations shown in Eq.~(\ref{eq:closed_form_cont_single}) can  actually be written by means of a covariance matrix as \cite{wang15}\footnote{Without loss of generality, we can assume that the input data and the shape displacements are zero-mean}:
\begin{equation}
\label{eq:covars_form}
Cov(X,Y) = Cov(X,X) \R,
\end{equation}
where we used the term $Cov$ to refer to the \textit{data} covariance, as opposite to $\Sig$ referring to the covariance of the sampling pdf. While \cite{levin02} applies a polytope approximation to estimate the covariance matrix, we rely on the first-order Taylor expansion of the feature space. We have to note that \cite{levin02} approximates the covariance matrix by considering combinations of training samples, while in our case we want to generate perturbations on the training data, and therefore the polytope approximation would not fit our problem. We can compute the functional covariance matrix as follows:
\begin{equation}
Cov(X,X) = \sum_{j} \int_{\delta \s} p(\delta \s) f(\I, \s_{j}^* + \delta \s) f(\I, \s_{j}^* + \delta \s)^T d \delta \s.
\end{equation}

If we approximate the input features by its first-order Taylor expansion, we can see that:
\begin{eqnarray}
 f(\I, \s_{j}^* + \delta \s) f(\I, \s_{j}^* + \delta \s)^T  \approx \nonumber \\
  \x_j^* \x_j^{*^T} + \x_j^* \delta \s^T \J_j^{*^T} + \J_j^* \delta \s \x_j^{*^T} + \J_j^* \delta \s \delta \s^T \J_j^{*^T} .
\end{eqnarray}

We can use Eq.~(\ref{eq:integrals2}), and the fact that $\int p(\delta \s) \delta \s \delta \s^T = \Sig + \bmu \bmu^T$ to further expand the covariance as:
\begin{equation}
Cov(X,X) = \sum_j \x_j^*{\x_j^*}^T+2\x_j^*\bmu^T{\J_j^*}^T + \J_j^*(\Sig+ \bmu\bmu^T){\J_j^*}^T,
\end{equation}
which is, exactly, the invertible part of Eq.~(\ref{eq:closed_form}). We can readily see that the covariance can be expressed as:
\begin{equation}
\label{eq:covars}
Cov(X,X) = \bar{\D}^* \hat{\B} (\bar{\D}^*)^T.
\end{equation}

Similarly, we can see that:
\begin{eqnarray}
\label{eq:covxy}
Cov(X,Y) = \sum_j \int_{\delta \s} \delta \s f(\I, \s_{j}^* + \delta \s)^T d \delta \s \approx \nonumber \\ 
\approx \sum_j  \bmu {\x_j^*}^T+ (\Sig + \bmu \bmu^T){\J_j^*}^T.
\end{eqnarray}

Obtaining Eq.~(\ref{eq:closed_form_cont}) from Eq.~(\ref{eq:covars_form}) is straightforward given  Eq.~(\ref{eq:covxy}) and Eq.~(\ref{eq:covars}). Interestingly, we can take advantage of this approach to generate our PCA models for each cascade level. These models are used to reduce the dimensionality of the feature space, and therefore need to be computed in the perturbed image space. This way, we can easily generate each of the models just by applying an eigendecomposition of the functional covariance of Eq.~(\ref{eq:covars}).

\subsection{Geometric interpretation}
\label{ssec:geometric}
\begin{figure}[t!]
\begin{center}
\includegraphics[width=0.95\columnwidth]{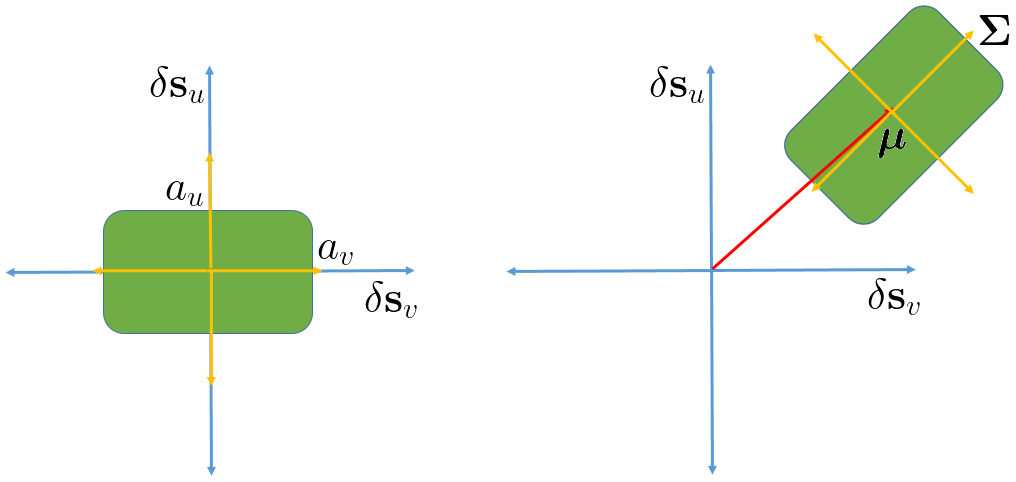}
\caption{Difference between  classical Functional Regression  (\textbf{Left}) and CR in correlated variables (\textbf{Right}). The  green area represents the volume within which samples are taken. Left: a diagonal covariance matrix, with entries defined as $\frac{a^2}{3}$. Right: a full covariance matrix and a non-zero mean vector. The sampling space is translated to the centre, defined by $\bmu$, and rotated according to the eigenvectors of $\Sig$.}
\label{f:differences}
\end{center}
\end{figure}

\noindent Finally, we want to elaborate on the meaning of the ``data term" in the Continuous Regression framework, i.e., the role of a pdf given that an \textit{infinite} set of samples is taken. To do so, we can analyse the solution presented in this paper from the Information Geometry perspective \cite{amari85,pennec06}, in which a geometrical interpretation can be applied to the probability measure. In the uncorrelated case, i.e. when using a diagonal covariance matrix, the manifold of shape displacements is a $2n$-dimensional parallelepiped, spanned by the Cartesian basis, defined for each of the shape dimensions. When we extend the Continuous Regression to full covariance matrices, we are rotating the manifold of shape displacements according to the eigenvectors of $\Sig$, and the origin of the coordinates axes is displaced to $\bmu$. Fig.~\ref{f:differences} illustrates this geometrical interpretation graphically.

\section{Incremental Cascaded Continuous Regression}
\noindent This section introduces the process of updating a CCR model using a new set of images and estimated shapes. We will show that our incremental CCR (which we coin iCCR) has a complexity that yields real-time performance. To the best of our knowledge, \textbf{the iCCR update is the first Cascaded Regression method achieving real-time capabilities}. Results shown in Section~\ref{sec:experimental_results} demonstrate the importance of incremental learning to achieve state-of-the-art results, thus illustrating the advantages of a real-time incremental learning algorithm.

\begin{figure}[t!]
\begin{center}
\includegraphics[width=0.95\columnwidth]{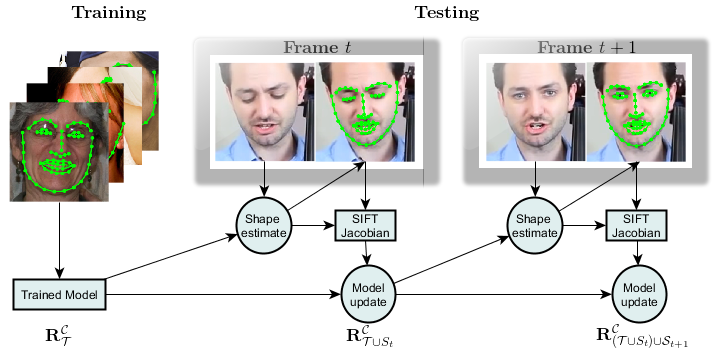}
\caption{Overview of our incremental cascaded continuous regression algorithm (iCCR). The originally model $\R_{\mathcal{T}}$ learnt offline is updated with each new frame, thus sequentially adapting to the target face.}
\label{fig:ilearn}
\end{center}
\end{figure}

Once a regressor has been trained, we might want to incorporate new images to the model, without the need of recomputing Eq.~(\ref{eq:closed_form_cont}). This is very important for the task of face tracking, since it has been reported that generic models perform worse than person-specific models. Since training a person-specific model a priori is not an option in most use cases, adding samples to a generic model as the face is being tracked may help the model to track a specific person better in future frames. An overview of the incremental learning procedure is depicted in Fig.~\ref{fig:ilearn}.

However, we can not retrain the full models by incorporating new images to the original training set, as it would be far too slow, and therefore is not computationally tractable. Some previous works \cite{xiong14,asthana14} have attempted to incorporate the online learning to their current models, by applying the recursive least squares to a trained model. However, both methods are very slow, and thus are impractical for real-time tracking. It should be noted that the good results reported by \cite{xiong14,asthana14} rely on the fact that videos were tracked offline, meaning that the update methods could take their time to learn while following frames politely waited for their turn to be tracked. Such affordances are not present when tracking from a live video stream, e.g. a webcam.

Let us devise the incremental learning update rule for the Continuous Regression. To do so, let us assume that we have a regressor $\R_\mathcal{T}$, trained using Eq.~(\ref{eq:compact_form_cont}), on a training set $\mathcal{T}$. Also, we denote the covariance matrix for the training data as $ \V_\mathcal{T} := Cov(X,X) = \bar{\D}_\mathcal{T}^* \hat{\B}(\bar{\D}_\mathcal{T}^*)^T$. The incremental (also known as online) learning aims to update $\R_{\mathcal{T}}$ with a set of $\mathcal{S}$ new images, for which the ground-truth image features and Jacobians are extracted. Let $\D_\mathcal{S}^*$ be the data corresponding to the updating samples. We can define a forgetting factor $\lambda$, which sets the influence of a given frame (the lower the $\lambda$, the more influence a frame has on the update), and define ${\bf W}_\lambda = \lambda \mathbb{I}_{2n+1}$, with $\mathbb{I}_{2n+1}$ the $(2n+1)$-dimensional matrix. Then, the new regressor would be computed as:
%\footnote{Sometimes, it is beneficial to include a \textit{learning factor} $\lambda$, which is used to set the weight of the new samples to be included. For the sake of clarity, we omit it in our derivation of incremental learning, although its inclusion would be straightforward} . We would compute the new regressor as:
\begin{equation}
\R_{\mathcal{T} \cup \mathcal{S}} = \M \left(  \sum_{j=1}^{M} \D_j^* + {\bf W}_\lambda^{-1} \D_\mathcal{S}^* \right)^T \left( \V_{\mathcal{T} \cup \mathcal{S}}\right)^{-1},
\end{equation}
where
\begin{equation}
\label{eq:new_covar}
\left( \V_{\mathcal{T} \cup \mathcal{S}} \right)^{-1} =  \left( \V_\mathcal{T} + \D_\mathcal{S}^* \B {\bf W}_\lambda^{-1} \D_\mathcal{S}^{*^T} \right)^{-1}.
\end{equation}
\noindent We can apply the Woodbury identity \cite{brookes11} to Eq.~(\ref{eq:new_covar}):
\begin{multline}
\label{eq:closed_form_icont}
{\V_{\mathcal{T} \cup \mathcal{S}}}^{-1} = {\V_{\mathcal{T}}}^{-1} - \\
 {\V_{\mathcal{T}}}^{-1} \D_{\mathcal{S}}^*\left( {\bf W}_\lambda \B^{-1}+{\D_{\mathcal{S}}^*}^T {\V_{\mathcal{T}}}^{-1} \D_{\mathcal{S}}^* \right)^{-1} {\D_{\mathcal{S}}^*}^T{\V_{\mathcal{T}}}^{-1}.
\end{multline}

This way, obtaining the inverse of the covariance matrix is computationally feasible. During tracking, the set $\mathcal{S}$ consists of a tracked image with its estimated landmarks, assuming these to have been correctly fitted. 
We can readily see that the sampling cost of the iCCR update is fixed to $\oO(5q)$, and does not depend on the number of cascade levels, whereas the incremental SDM (iSDM), requires the sampling process to be carried out for each cascade level. We can see that Eq.~(\ref{eq:closed_form_icont}) needs to compute the inverse of the matrix $ {\bf W}_\lambda \B^{-1}+{\D_{\mathcal{S}}^*}^T {\V_{\mathcal{T}}}^{-1} \D_{\mathcal{S}}^*$, which is $(2n+1)$-dimensional. This operation has a complexity of $\oO( (2n+1)^3)$. That is, the computational complexity of inverting that matrix is cubic with respect to the number of points. However, we can alleviate this computation by working with a PDM \cite{cootes04}, instead of predicting over the points in a direct way. In a PDM, a shape $\s$ is parameterised in terms of $\p = [\q,\c] \in \Real^{m}$, where $\q\in\Real^{4}$ represents the rigid parameters and $\c$ represents the flexible shape parameters, so that $\s = t_\q( \s_0 + \B_s \c )$, where $t$ is a Procrustes transformation parameterised by $\q$. $\B_s \in \Real^{ 2n \times m}$ and $\s_0\in \Real^{2n}$ are learnt during training and represent the linear subspace of flexible shape variations. This way, the Continuous Regression formulation would be simply transformed into:
\begin{equation}
\label{eq:parametrised_continuous}
\sum_{j=1}^M \int_{\delta \p} p( \delta \p ) \| \delta \p - \R f( \I_j, \p^*_j+\delta \p ) \|_2^2 d \delta \p,
\end{equation}
where the Jacobians now need to be computed with respect to the PDM. To do so, it suffices to apply the chain rule:
\begin{equation}
\frac{\partial f( \I,\p + \delta \p)}{\partial \p} = \frac{\partial  f( \I,\p + \delta \p)}{\partial \s}\frac{ \partial \s}{\partial \p} ,
\end{equation}
where $\frac{ \partial \s}{\partial \p}$ can be analytically computed from the shape model, and $\frac{\partial  f( \I,\p + \delta \p)}{\partial \s}$ is computed as in Eq.~(\ref{eq:derivative}). This way, the Continuous Regression solution is exactly the same as shown in Eq.~(\ref{eq:closed_form_cont}), with the Jacobians now being computed with respect to the PDM. 
Now, the matrix ${\bf W}_\lambda \B^{-1}+{\D_{\mathcal{S}}^*}^T {\V_{\mathcal{T}}}^{-1} \D_{\mathcal{S}}^*$ is $m$ dimensional, and thus its inversion is $\oO(m^3)$. In our experimental set-up, $2n = 132$, and $m = 24$. The computational saving is obvious. 

Now, we can further analyse the update complexity, noting that the most expensive step in the updating process is not computing the inverse of $ {\bf W}_\lambda \B^{-1}+{\D_{\mathcal{S}}^*}^T {\V_{\mathcal{T}}}^{-1} \D_{\mathcal{S}}^*$, but rather computing ${\D_{\mathcal{S}}}^T{\V^{\mathcal{C}}_{\mathcal{T}}}^{-1}$, which has a complexity cost of $\mathcal{O}(d^2m)$ (assuming a PDM is used). The overall computational cost of updating a regressor is
\begin{equation}
\label{eq:cost_cont}
\oO( 3m d^2) + \oO(3m^2 d) + \oO(m^3).
\end{equation}

We compare this updating cost with the \textbf{sampling-based} SDM update proposed in \cite{asthana14}, which was formulated as:
\begin{eqnarray}
\R_{\T \cup \mathcal{S}} & = & \R_\T - \R_\T \Q + \Y_{\mathcal{S}}\X_{\mathcal{S}}^T\V_{\mathcal{T}\cup \mathcal{S}} \nonumber \\
\Q & = & \X_{\mathcal{S}} \U \X_{\mathcal{S}}^T\V_{\mathcal{T}} \nonumber \\
\U & = & \left( \mathbb{I}_K + \X_{\mathcal{S}}^T \V_{\mathcal{T}} \X_{\mathcal{S}} \right)^{-1}  \nonumber \\
\V_{\mathcal{T} \cup \mathcal{S}} & = & \V_{\mathcal{T}}- \V_{\mathcal{T}}\Q 
\end{eqnarray}
\noindent where $\mathbb{I}_K$ is the $K$-dimensional identity matrix. We can readily see that the main bottleneck in this update comes from computing $\V_{\mathcal{T}}\Q$, which is $\oO( d^3 )$, i.e., the \textit{iCCR update is an order of magnitude faster than the sampling-based SDM update} proposed in \cite{asthana14}. This brings the cost of updating from the $\sim 4$ seconds reported in \cite{asthana14} down to $\sim 0.2$ seconds, which is the total update time in our sequential Matlab implementation of iCCR.

%%%

\section{Experimental results}
\label{sec:experimental_results}
\noindent To evaluate our proposed approach, we have tested our CCR and iCCR methods on the most extensive benchmark that exists to date:  300VW \cite{shen15}. To compare our proposed methods against top participants, we develop a fully automated system, which includes automated initialisation, as well as a  tool to detect that tracking is lost. This way, we can compare our methods in exactly the same conditions that were set up for the benchmark. We will show that our fully automated system achieves state of the art results, as well as the importance of incremental learning. Code for training and testing is available on the lead author's website. 

\subsection{Experimental set-up}
\label{ssec:Experimental_set_up}
\noindent \textbf{Test Data:} All methods are tested on the 300VW\cite{shen15} dataset, which, to the best of our knowledge, is the most extensive and recent benchmark in Face Tracking. The dataset consists of 114 videos, each $\sim 1$ minute long, divided into different categories. 50 videos are used for training, whereas the 64 remaining videos are subdivided into three categories, intended to represent increasingly unconstrained scenarios. Category 1 contains 31 videos recorded in controlled conditions, whereas Category 2 includes 19 videos recorded under severe changes in illumination. Category 3 contains 14 videos captured in totally unconstrained scenarios. All the videos are of a single-person, and have been annotated in a semi-supervised manner using two different methods \cite{tzimiropoulos15,chrysos15}. All frames in which the face appears beyond profile-view have been removed from the challenge evaluation, and therefore are not considered in our experiments either.

\noindent \textbf{Error measure:} The error measure is that defined for 300VW. It is computed for each frame by dividing the average point-to-point Euclidean error by the inter-ocular distance, understood to be the distance between the two outer eye corners. More specifically, if $\hat{\s}$ is the estimated shape, and $\s^*$ is the ground-truth shape, the RMSE is given as:
\begin{equation}
RMSE = \frac{\sum_{i=1}^n\sqrt{(\hat{x}_i - x^*_i)^2 + (\hat{y}_i - y^*_i)^2}}{d_{outer}n},
\end{equation}
where $d_{outer}$ is the Euclidean distance between the points defined for the outer corner of the eyes, measured on the ground-truth. The results are summarised in the form of Cumulative Error Distribution curves (CED), along with the Area Under the Curve (AUC). 

\subsection{Training}
\noindent \textbf{Data:} We use data from different datasets of static images to construct our training set. Specifically, we use a total of $\sim$7000 images taken from Helen \cite{le12}, LFPW \cite{belhumeur11}, AFW \cite{zhu12}, IBUG \cite{sagonas13}, and a subset of MultiPIE \cite{gross10}. We have used the facial landmark annotations provided by the 300 faces in the wild challenge \cite{shen15,sagonas13b}, as they are consistent across datasets. Our models are trained for a 66-point configuration. In order to ensure the consistency of the annotations with respect to the test set, the Shape Model is constructed using the training partition of 300VW, and has $20$ non-rigid parameters and $4$ rigid parameters \cite{matthews04}. 

\noindent \textbf{Training:} As shown in Algorithm~\ref{alg:ccr_train}, the training of CCR starts from a set of given statistics ($\bmu^0$ and $\Sig^0$). In our experiments, the statistics are computed across the training sequences, by computing the differences between consecutive frames. That is to say, $\bmu^0$ and $\Sig^0$ are meant to model how the shapes vary from frame to frame. The main idea of this approach is to replicate a real scenario when ``perturbing" the ground-truth. Given that the training set shows smoother variations w.r.t. the test set, we compute the differences between frames separated by two and three time steps. This way, higher displacements are also captured.

\noindent \textbf{Features:} We use a HOG \cite{dalal05} implementation, with a block-size of $24$ pixels, subdivided in $4$ blocks, and $9$ bins, resulting in a $9504$-dimensional vector. We apply PCA on the output, retaining 2000 dimensions, i.e., $d = 2000$. To generate a PCA model per cascade level, we compute the functional covariance shown in Eq.~(\ref{eq:covars}). 

\noindent \textbf{Cascade levels:} In our experimental set-up, we have fixed the number of Cascade levels to $L = 4$. 

\subsection{Testing}
\label{ssec:testing_set_up}
\noindent \textbf{Initialisation:} The tracker is initialised at the beginning of each sequence, as well as each time the tracker is detected to have lost a fitting (see details below). The initialisation utilises the open-source dlib face detection to locate the face bounding box (\url{dlib.net}), and then predicts the shape with a Context-based SDM \cite{sanchez16}. Then, a single CCR iteration is carried out.  \\

\noindent \textbf{Incremental learning update:} The incremental learning needs to filter frames to decide whether a fitting is suitable or harmful to be used for the model updates.  We use a simple heuristic with a linear SVM, trained to decide whether a particular fitting is ``correct", understood as being below a threshold error. In order to avoid redundancy, once a frame has been used to update the models, we consider no more frames within a window of 3 consecutive frames. Given that videos are of a single person, we do not need to perform any face recognition to guarantee that updated models keep tracking the same person with updated, and thus personalised, models. For the iCCR update, we collect person-specific statistics on a sliding window of 50 frames, so that the person-specific variations are encoded in the statistics used to update the model. The forgetting factor $\lambda$, has been found through validation on the training set, and is set to $\lambda^1 = 0.01$, $\lambda^2 = 0.025$, $\lambda^3 = 0.05$ and $\lambda^4 = 0.1$, where $\lambda^l$ is the learning rate for the $l$-th cascade level, and the lower the $\lambda$, the more influence a frame has on the update. We evaluate the sensitiveness of iCCR w.r.t. $\lambda$ in Section~\ref{ssec:lambdas}. 

\noindent \textbf{Loss of tracking detection:} Sometimes the tracker fails to track a face. This occurs in particular when there are significant occlusions or head-poses beyond profile. In these cases, we need to detect that the tracker has lost a face, in order to reinitialise it in the next frame from the face detector output. We used the same SVM learnt for the incremental learning update to also detect whether the tracker is lost or not, by empirically selecting a suitable threshold for the score returned by the SVM model.

\subsection{Influence of the forgetting factor}
\label{ssec:lambdas}
\begin{figure}[t!]
\begin{center}
\includegraphics[width=1\columnwidth]{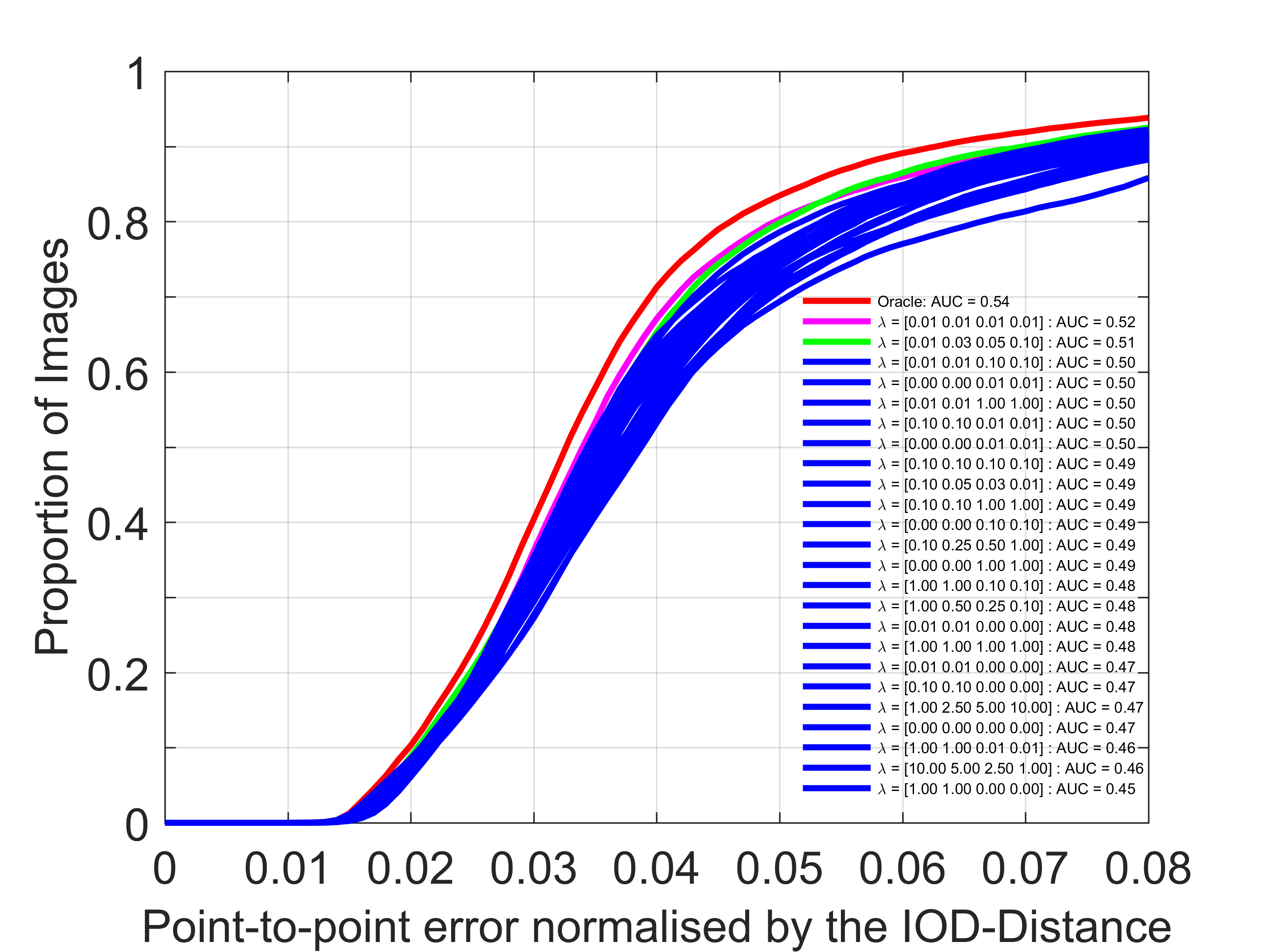}
\caption{CED's for a set of different forgetting factors, along with the corresponding AUCs (Category 3). Red curve shows the Oracle. Magenta shows the best factor found for that Category. Green curve shows the results attained by the forgetting factor found on the training set. Blue curves are results attained by other learning rates \label{fig:lambdas} }
\end{center}
\end{figure}
\noindent The forgetting factor has been found through validation on the training partition, and set up as shown above. In general, it has been found that the top levels of the cascade are less sensitive to an error in the update, and therefore a smaller factor can be used (meaning a higher impact of the given frame). However, for the last levels of the cascade, a low factor can have a big impact if the frame to be updated is not ``entirely" correct. Since the score of an SVM is used to detect whether a frame is suitable or not for the update, it is likely that eventually a partially correct frame will be used. To measure how well the found values generalise to the test set, we have included an experiment in Category 3 showing the performance of other forgetting factors. The overall results are shown in Fig.~\ref{fig:lambdas}. It can be seen that other values would produce similar results. In addition, we include the error given by an ``Oracle'' (red curve), understood as the aggregated error for a scenario in which the best forgetting factor is chosen for each video independently. That is to say, red curve represents the results that would be attained if we could choose the best forgetting factor per video (fixed for the whole video). It can be seen that the red curve does not show a significant improvement over the best blue curve, meaning that in general the learning rate might have a limited impact in the results. 

\subsection{Equivalence with SDM}
\label{ssec:equivalence}
\noindent In order to demonstrate the performance of our CCR method compared to SDM, we have trained an SDM model under the same conditions as for CCR. More specifically, we have used the same training set, and the same statistics measured for the first level. The training was then done in a sequential manner, in a similar fashion to that of Algorithm~\ref{alg:ccr_train}. The number of cascade levels was again set to 4. The PCA models were directly computed over the sampled images, and comprise 2000 components. The SDM model was tested in the test partition of 300-VW under exactly the same conditions than those of CCR and iCCR. The results attained by our SDM model are shown in Fig.~\ref{fig:soacomp}.

\subsection{Comparison with state of the art}
\label{ssec:comparisons_with_sota}
\begin{figure}[h!]
\begin{center}
\includegraphics[width=0.9\columnwidth]{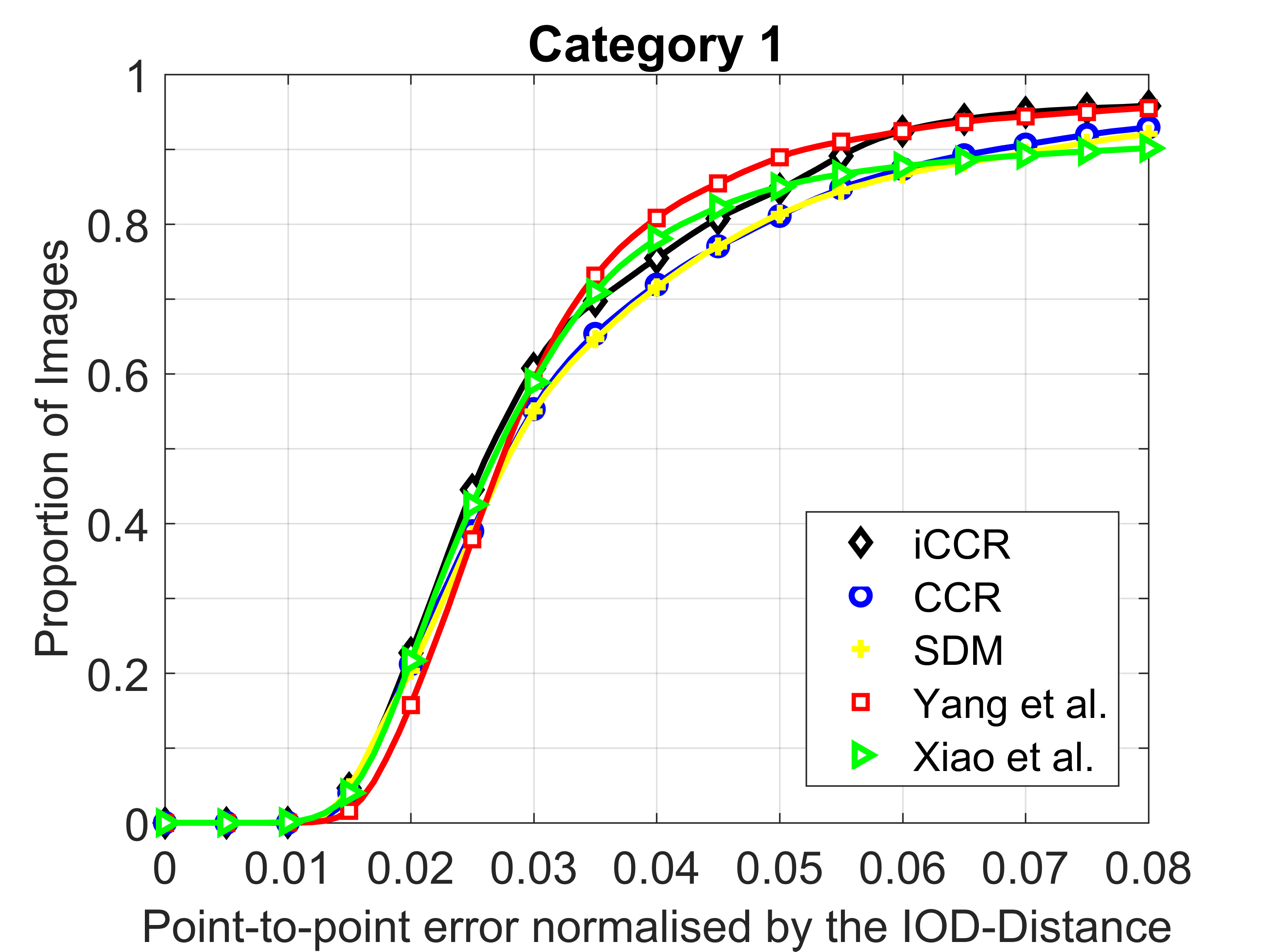} \\
\includegraphics[width=0.9\columnwidth]{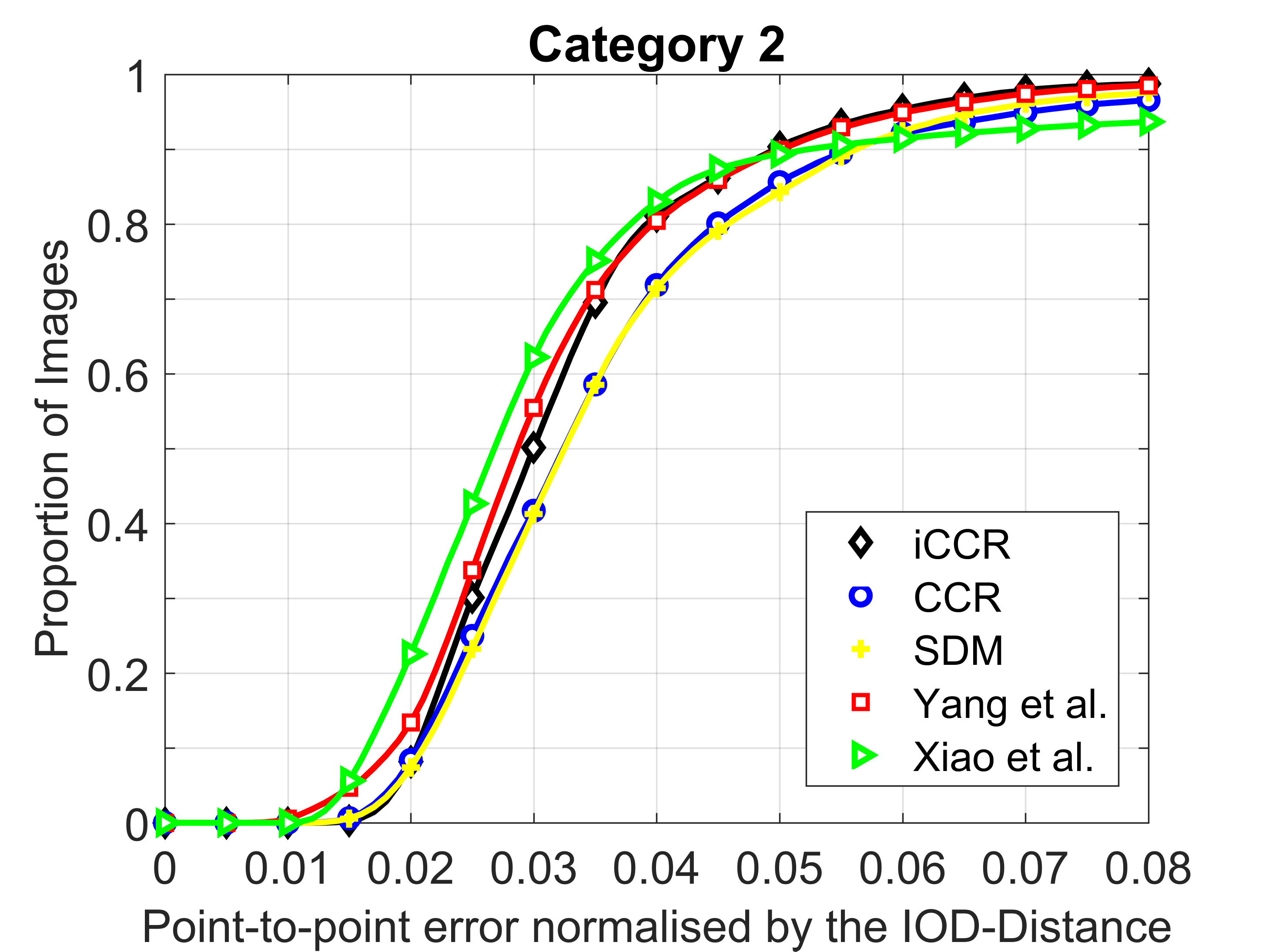} \\
\includegraphics[width=0.9\columnwidth]{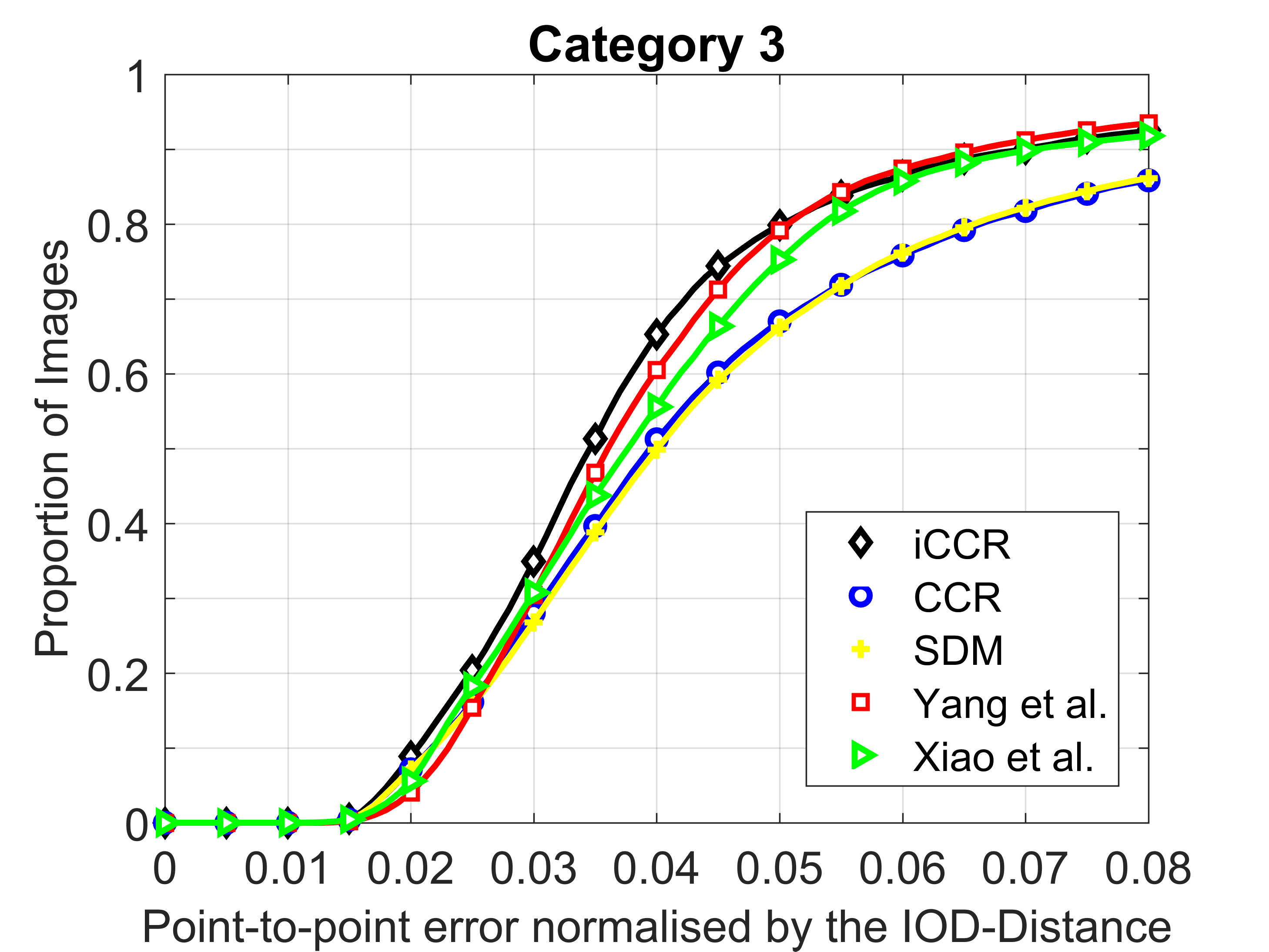} \\
\caption{CED's for the 49-points configuration. Better seen in colour \label{fig:soacomp} }
\end{center}
\end{figure}

\noindent We evaluate our fully automated tracking system in exactly the same conditions as those used for the 300VW benchmark. We compare our method with respect to the top two 300VW results \cite{yang15,xiao15}. We report results for the configurations of 49 points. In order to compare our method against \cite{yang15,xiao15}, we asked these authors for their results, which were kindly provided. For the fairest comparison with respect to the 300VW evaluation, we removed the frames that were not considered for the evaluation (i.e. those including faces beyond profile). To show the benefits of incremental learning,  we include the results of both our CCR and iCCR systems in all curves, as well as our trained SDM. CED results are shown in Fig.~\ref{fig:soacomp}, and the AUC for each method is shown in Table~\ref{t:res}. It is interesting to remark that the methods we compare with made contributions to the initialisation of the tracker in each frame, or to failure detection. In our case, even with a simple heuristic for both initialisation and failure detection, we are capable of attaining comparable or even superior performance over \cite{yang15,xiao15}. The initialisation for each frame in our tracking system is simply the output of the previous frame. We do not include any multi-view models or progressive initialisation, but instead ensure that the statistics used to generate our models generalise well to unseen scenarios. 

It is unlikely that algorithms with slower running times than iCCR, such as \cite{yang15,xiao15}, would attain their high accuracies if forced to process challenging videos in real-time. This has important consequences for tracking streaming video, in which frames must be tracked as fast as they arrive in order to make use of the limited movement of facial points between adjacent frames, something that is beneficial to trackers of any ilk.

\begin{figure*}[t!]
\begin{center}
\includegraphics[width=1\columnwidth]{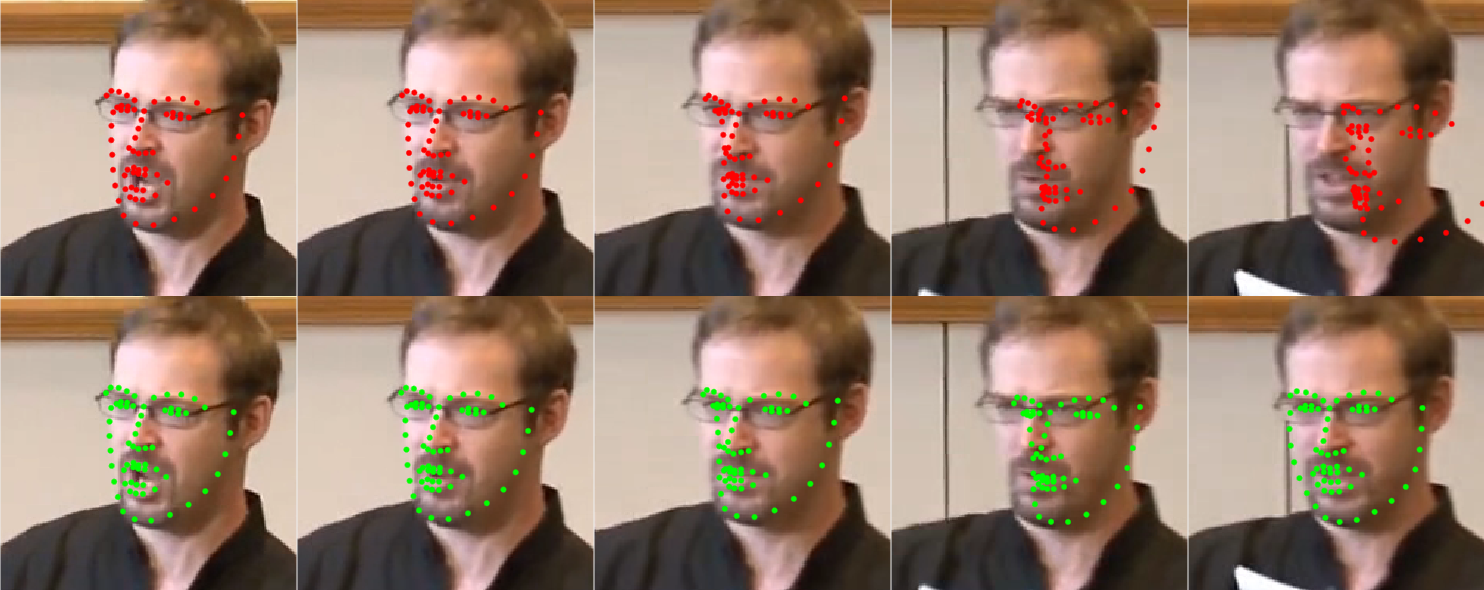} 
\includegraphics[width=1\columnwidth]{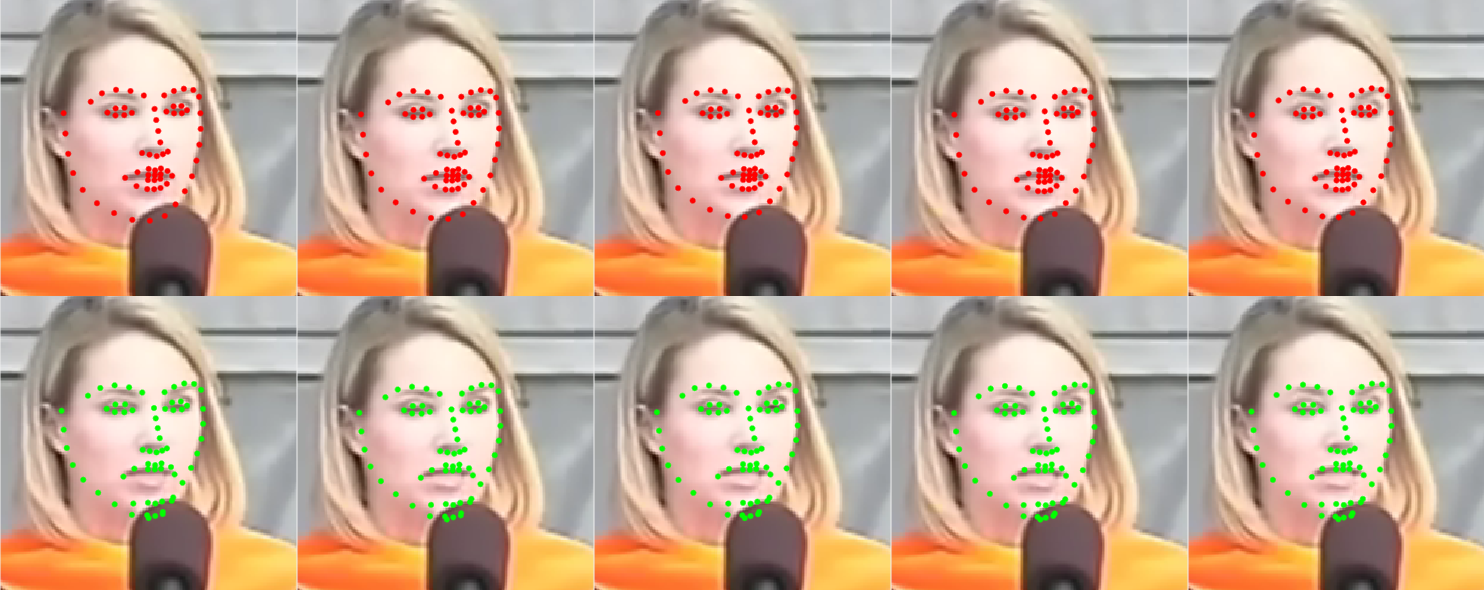} \\
\includegraphics[width=1\columnwidth]{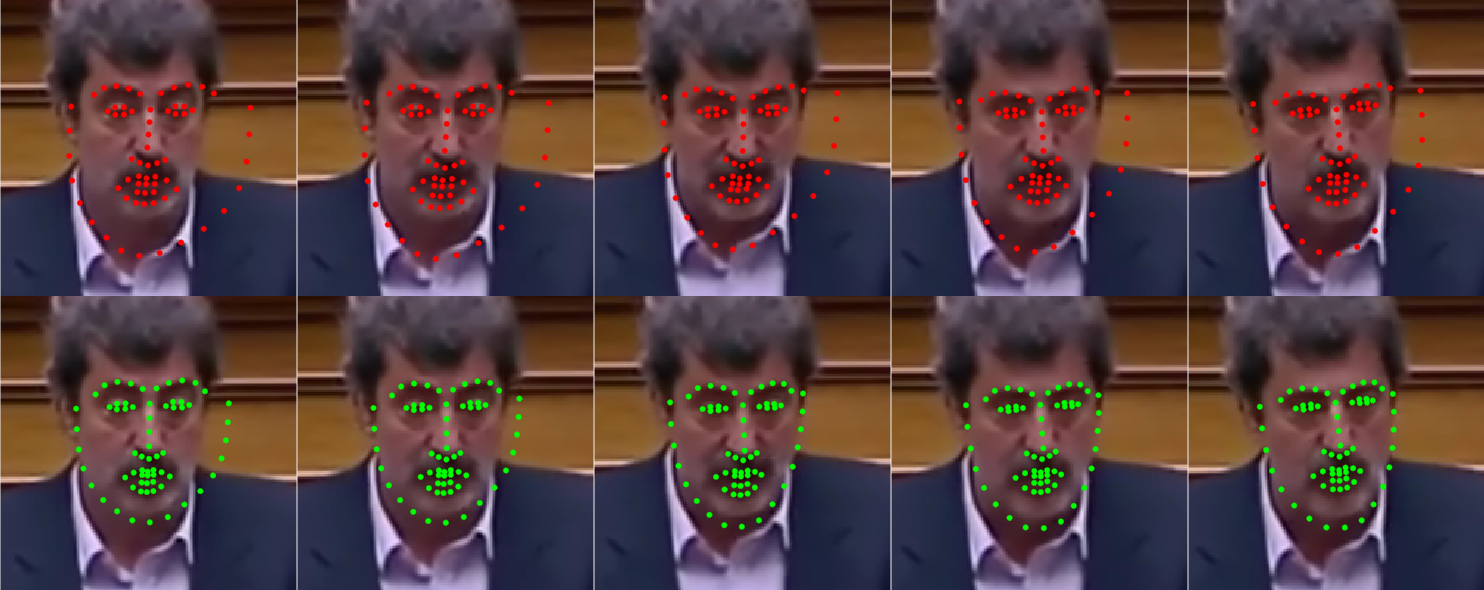} 
\includegraphics[width=1\columnwidth]{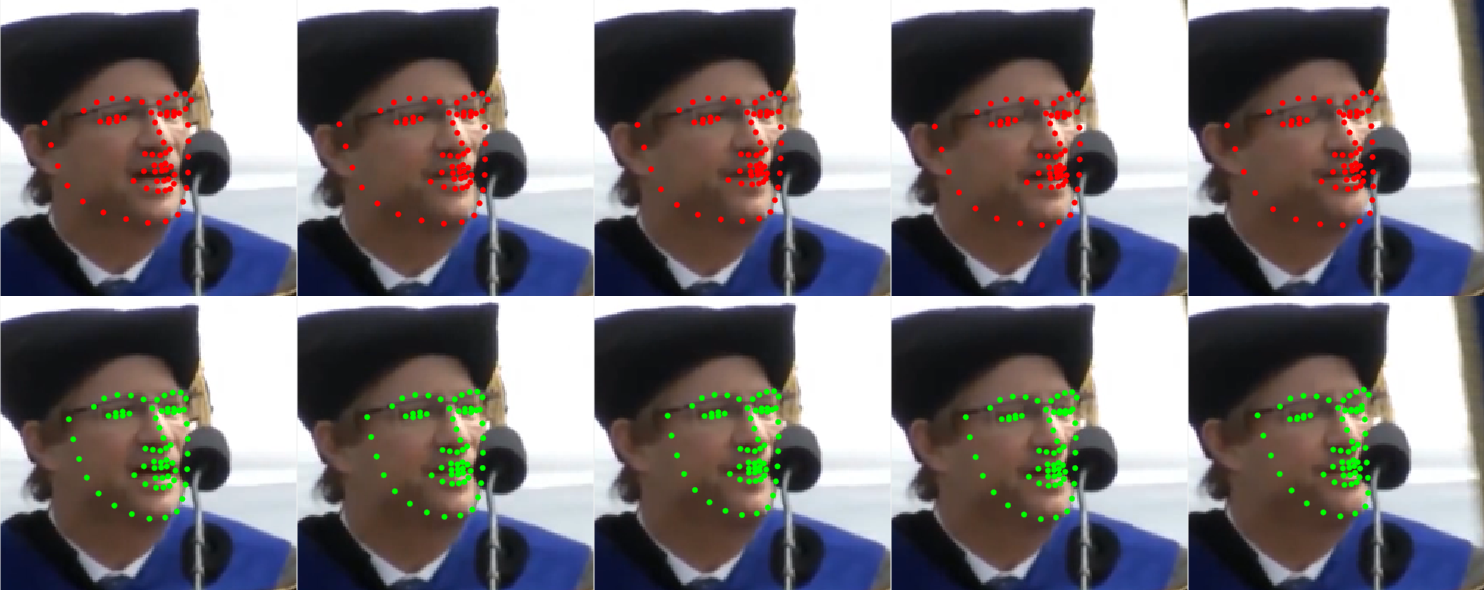} \\
\caption{Visual examples from Category 3 on 300VW. Upper rows show the CCR fitting (red points), whereas the lower rows show the iCCR fitting (green points). The sequences on the left show a clear case where iCCR outperforms CCR thanks to its incremental learning capabilities (fast movement in the upper sequence, low resolution in the sequence at the bottom). The sequences on the right show a failure case where the learned features are not capable to outperform CCR. All qualitative videos and results can be found at \protect \url{https://continuousregression.wordpress.com/code}. \label{fig:examples} }
\end{center}
\end{figure*}

It is worth highlighting that CCR and SDM models were reinitialised in $\sim 0.51\%$ and $\sim 0.56\%$ of total frames ($>121000$), respectively, whilst the iCCR was reinitialised in only $\sim 0.34\%$ of the frames. This illustrates the stability of our models, resulting in a simple yet effective tracking system. Specially remarkable is the importance of the Incremental Learning in challenging sequences, such as those shown in Category 3, to achieve state of the art results. As shown throughout the paper, the complexity of our incremental learning allows for real-time implementation, something that could not be achieved by previous works on Cascaded Regression. Some visual examples are depicted in Fig.~\ref{fig:examples}. Code and videos can be found at \url{http://continuousregression.wordpress.com}.

\begin{table}[h!]
\begin{center}
  \begin{tabular}{ | c || c | c | c | c | c | c | }
    \hline
    Method & Category 1 & Category 2 & Category 3\\ \hline
   iCCR  &  0.5978 & 0.5918 & \textbf{0.5141}  \\ \hline
   CCR  & 0.5657 & 0.5539 & 0.4410 \\  
    \hline
    SDM  & 0.5617 & 0.5522 & 0.4380 \\  
    \hline
Yang et al. \cite{yang15} & \textbf{0.5981} &  0.6025 & 0.4996    \\ \hline
 Xiao et al.  \cite{xiao15} & 0.5814 & \textbf{0.6093} & 0.4865 \\ \hline
  \end{tabular}
\end{center}
\caption{\label{t:res} AUC for 49 points configuration for the different categories. }
\end{table}  

\section{Conclusion}
\noindent We have proposed a novel formulation for the problem of Continuous Least Squares in spaces of correlated variables, namely Continuous Regression. We have shown the relation that exists between the sampling distribution and the space within which samples are taken, and have proposed a solution for general feature descriptors. We then incorporated the Continuous Regression within the Cascaded Regression framework, and showed its performance is similar to sampling-based methods. We devised an incremental learning approach using Continuous Regression, and showed its complexity allows for real-time performance. Results demonstrate the state-of-the-art performance in challenging sequences, whilst being able to run in real-time. We also want to note that while we tackle the facial landmark tracking problem, cascaded regression has also been applied to a wider range of problems such as pose estimation \cite{dollar10}, model-free tracking \cite{wang15b} or object localisation \cite{yan14}, thus meaning the contributions of this paper can be extended to other problems. Future work will include also the optimisation of the learning process, given that a better classifier than a SVM would probably select better frames, and we will also investigate the effect of the forgetting factor under different configurations.

\ifCLASSOPTIONcompsoc
  \section*{Acknowledgments}
\else
  \section*{Acknowledgment}
\fi

\noindent The work of S\'anchez-Lozano, Martinez and Valstar was supported by the European 
Union Horizon 2020 research and innovation programme under grant agreement No
645378, ARIA-VALUSPA. The work of S\'anchez-Lozano was also supported by the 
Vice-Chancellors Scholarship for Research Excellence of the University of
Nottingham. The work of Tzimiropoulos was supported in part by the EPSRC project
EP/M02153X/1 Facial Deformable Models of Animals. We are also grateful for access to the University of Nottingham High Performance Computing Facility.

\ifCLASSOPTIONcaptionsoff
  \newpage
\fi



%\bibliographystyle{IEEEtran}

\begin{IEEEbiography}[{\includegraphics[width=1in,height=1.25in,clip,keepaspectratio]{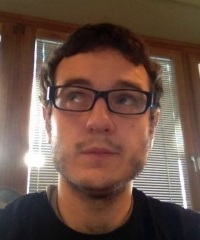}}]{Enrique S\'anchez-Lozano} is a Research Fellow in the Computer Vision Lab at the University of Nottingham, where he received his PhD in Computer Science (2017). He received his M.Eng. in Telecommunication Engineering (2009) and M.Sc. in Signal Theory and Communications (2011) from the University of Vigo (Spain). He spent 6 months as a visiting scholar at the Human Sensing Lab, with Fernando De la Torre, between 2011 and 2012. He was given a Fundaci\'on Pedro Barri\'e Scholarship to do a 5 month research stay at the Computer Vision Lab (CVL), University of Nottingham, with Michel Valstar, between 2013 and 2014.
\end{IEEEbiography}
\vspace{-0.89cm}
\begin{IEEEbiography}
[{\includegraphics[width=1in,height=1.25in,clip,keepaspectratio]{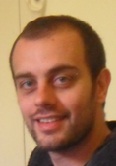}}]{Georgios (Yorgos) Tzimiropoulos}
received the
M.Sc. and Ph.D. degrees in Signal Processing and
Computer Vision from Imperial College London,
U.K. He is Assistant Professor with the School of
Computer Science at the University of Nottingham,
U.K. Prior to this, he was Assistant Professor with
the University of Lincoln, U.K. and Senior Researcher
in the iBUG group, Department of Computing,
Imperial College London. He is currently
Associate Editor of the Image and Vision Computing
Journal. His main research interests are in the areas
of face and object recognition, alignment and tracking, and facial expression
analysis.
\end{IEEEbiography}
\vspace{-0.89cm}
\begin{IEEEbiography}[{\includegraphics[width=1in,height=1.25in,clip,keepaspectratio]{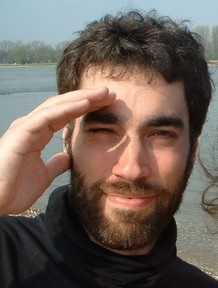}}]{Brais Martinez} (M'10) is a Research Fellow with the Computer Vision Laboratory at the University of Nottingham, UK. He has previously obtained his PhD in Computer Science from the Universitat Autonoma de Barcelona, and held a postdoctoral research position at the Imperial College London. 
He has worked on the topics of Computer Vision and Pattern Recognition, mainly focusing on model-free tracking and on face analysis problems such as face alignment, face detection and facial expression recognition, publishing his findings on these topics on a variety of authoritative venues such as CVPR, ICCV, ECCV or TPAMI.
\end{IEEEbiography}
\vspace{-0.89cm}


%
%% insert where needed to balance the two columns on the last page with
%%\newpage
%
\begin{IEEEbiography}[{\includegraphics[width=1in,height=1.25in,clip,keepaspectratio]{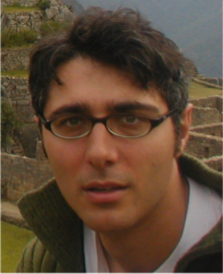}}]{Fernando De la Torre} is an Associate Research
Professor in the Robotics Institute at Carnegie
Mellon University. He received his B.Sc. degree
in Telecommunications, as well as his M.Sc. and
Ph. D degrees in Electronic Engineering from
La Salle School of Engineering at Ramon Llull
University, Barcelona, Spain in 1994, 1996, and
2002, respectively. His research interests are
in the fields of Computer Vision and Machine
Learning. Currently, he is directing the Component
Analysis Laboratory (http://ca.cs.cmu.edu)
and the Human Sensing Laboratory (http://humansensing.cs.cmu.edu)
at Carnegie Mellon University. He has over 150 publications in referred
journals and conferences and is Associate Editor at IEEE TPAMI. He has
organized and co-organized several workshops and has given tutorials
at international conferences on component analysis.
\end{IEEEbiography}
\vspace{-0.89cm}
\begin{IEEEbiography}[{\includegraphics[width=1in,height=1.25in,clip,keepaspectratio]{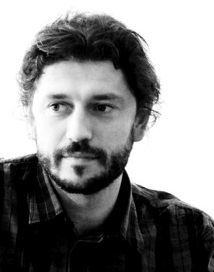}}]{Michel Valstar}
(M'09) is an Associate Professor in the Computer Vision and Mixed
Reality Labs at the University of Nottingham. He
received his masters degree in Electrical Engineering
at Delft University of Technology in 2005 and
his PhD in computer science with the intelligent
Behaviour Understanding Group (iBUG) at Imperial
College London in 2008. His main interest is in automatic recognition of
human behaviour. In 2011 he was the main organiser
of the first facial expression recognition challenge,
FERA 2011. In 2007 he won the BCS British Machine Intelligence Prize
for part of his PhD work. He has published technical papers at authoritative
conferences including CVPR, ICCV and SMC-B and his work has received
popular press coverage in New Scientist and on BBC Radio. 
\end{IEEEbiography}

%\vfill


\end{document}